\documentclass{article} 
\usepackage[final]{colm2025_conference}

\usepackage{microtype}
\usepackage{hyperref}
\usepackage{url}
\usepackage{booktabs}
\usepackage{graphicx}
\usepackage{wrapfig}
\usepackage{colortbl}
\usepackage{xcolor}
\usepackage{adjustbox}
\usepackage{multirow}
\usepackage{enumitem}
\usepackage{subcaption}
\usepackage{booktabs}
\usepackage{amsmath}
\usepackage{amsfonts}

\definecolor{lightblue}{rgb}{0.85,0.92,1} 
\definecolor{lightyellow}{rgb}{1, 0.95, 0.8}

\usepackage{lineno}
\usepackage{amssymb}

\definecolor{darkblue}{rgb}{0, 0, 0.5}
\hypersetup{colorlinks=true, citecolor=darkblue, linkcolor=darkblue, urlcolor=darkblue}
\usepackage[hang,flushmargin]{footmisc} 

\title{Enhancing LLM Reasoning with Iterative DPO: \\ A Comprehensive Empirical Investigation}

\author{\textbf{Songjun Tu$^\spadesuit$$^\clubsuit$$^\blacklozenge$},
\textbf{Jiahao Lin$^\blacklozenge$$^\spadesuit$},
\textbf{Xiangyu Tian$^\blacklozenge$$^\spadesuit$},
\textbf{Qichao Zhang$^\spadesuit$$^\blacklozenge$},
\textbf{Linjing Li$^\spadesuit$$^\blacklozenge$},\\
\textbf{Yuqian Fu$^\spadesuit$$^\blacklozenge$},
\textbf{Nan Xu$^\heartsuit$},
\textbf{Wei He$^\diamondsuit$},
\textbf{Xiangyuan Lan$^\clubsuit$},
\textbf{Dongmei Jiang$^\clubsuit$},
\textbf{Dongbin Zhao$^\spadesuit$$^\clubsuit$$^\blacklozenge$}
\\
 $^\spadesuit$~Institute of Automation, Chinese Academy of Sciences\quad
 $^\clubsuit$~Pengcheng Laboratory  \\
 $^\blacklozenge$~School of Artificial Intelligence, University of Chinese Academy of Sciences \\
 $^\heartsuit$~Wenge Technology\quad
 $^\diamondsuit$~Fudan University
\\
\texttt{\{tusongjun2023,zhangqichao2014\}@ia.ac.cn}
}

\begin{document}

\ifcolmsubmission
\linenumbers
\fi

\maketitle
\begin{abstract}
Recent advancements in post-training methodologies for large language models (LLMs) have highlighted reinforcement learning (RL) as a critical component for enhancing reasoning.
However, the substantial computational costs associated with RL-based approaches have led to growing interest in alternative paradigms, such as Direct Preference Optimization (DPO). 
In this study, we investigate the effectiveness of DPO in facilitating self-improvement for LLMs through iterative preference-based learning. 
We demonstrate that a single round of DPO with coarse filtering significantly enhances mathematical reasoning performance, particularly for strong base model. 
Furthermore, we design an iterative enhancement framework for both the generator and the reward model (RM), enabling their mutual improvement through online interaction across multiple rounds of DPO.
Finally, with simple verifiable rewards, our model DPO-VP achieves RL-level performance with significantly lower computational overhead.
These findings highlight DPO as a scalable and cost-effective alternative to RL, offering a practical solution for enhancing LLM reasoning in resource-constrained situations.
Code available at:  \href{https://github.com/TU2021/DPO-VP}{https://github.com/TU2021/DPO-VP}.
\end{abstract}

\renewcommand{\thefootnote}{}

\footnotetext{$*$ Corresponding authors: Qichao Zhang and Xiangyuan Lan.}

\section{Introduction}
The field of large language models (LLMs) is evolving at an unprecedented pace \citep{hurst2024gpt, xi2025rise}. 
With the release of O1 \citep{jaech2024openai} by OpenAI in September 2024, the research focus shifted from pre-training to post-training.
This transition was further accelerated in January 2025, when DeepSeek-R1 \citep{guo2025deepseek} brought post-training methodologies to the forefront of LLM research.
Then, several studies have explored reinforcement learning (RL) \citep{gandhi2025cognitive,zeng2025simplerl,tu2025online} at scale and supervised fine-tuning (SFT) \citep{muennighoff2025s1,ye2025limo, fu2025srft} with high-quality long-chain data, unlocking emergent capabilities such as self-checking, reflection, and verification in strong base models (e.g., Qwen2.5 \citep{yang2024qwen25}). 
These advancements highlight the potential of post-training in enhancing LLM reasoning.

Generally, searching self-generated data for training within post-training is referred to self-improvement \citep{yang2024qwen2}. 
At its core, this approach exploits search and learning to leverage the power of general methods as computational resources increase — resonating with the principle articulated by \citet{sutton2019bitter} in \textit{The Bitter Lesson}.
Naturally, search and expansion can be achieved through Monte Carlo Tree Search (MCTS) and RL. MCTS-based test-time scaling and reward model (RM) training were once considered key contributors behind O1 \citep{zhang2024rest, cui2025process}. But the high inference cost and training complexity gradually pushed MCTS out of the research spotlight.
In contrast, RL with verifiable rewards has been empirically shown to be more efficient and promising \citep{guo2025deepseek,xie2025logic}. 
However, current RL training still requires substantial computational resources for reproduction. 
For instance, SimpleRL \citep{zeng2025simplerl} trains for 1.5 days on 4×8 H100 GPUs, while PURE \citep{cheng2025pure} also demands 8 A100 GPUs. 
Achieving comparable or even superior self-improvement performance with lower computational resources and reduced training time is an ideal goal pursued in this field.

Several methods have been introduced as alternatives to RL to alleviate computational overhead, such as DPO \citep{rafailov2023direct}, KTO \citep{ethayarajh2024kto}, and IPO \citep{azar2024general}. 
By modeling implicit rewards, these approaches remove the dependency on explicit reward and value functions of RL and are collectively known as offline maximum likelihood estimation (MLE) methods. 
Through multi-round iterative training, offline MLE methods can be approximated as off-policy online RL approaches \citep{pang2024iterative}.
Recently, \citet{swamy2025all} demonstrated the theoretical equivalence between online RL and offline MLE under ideal optimization conditions.
While the role of RL in post-training has been extensively studied, the impact of the more streamlined and efficient DPO in post-training remains uncertain.

In this paper, we focus on empirically investigating the improvements brought by DPO in post-training, aiming to provide the LLM community with a deeper performance analysis of offline MLE self-improvement paradigms.
Our key findings are as follows:
\begin{enumerate}[leftmargin=*]
    \item \textbf{Single-Round DPO with Coarse Filtering Can Enhance Strong Base Models}: With simple RM or even outcome labels to filter self-generated datasets, we achieve a significant improvement in mathematical reasoning of Qwen2.5 with single-round DPO training.
    \item \textbf{Iterative Improvement of Generator and RM can be Achieved via Multi-Round DPO}: By relabeling the RM with online data, the mutual enhancement of the generator and RM can be realized through multiple rounds of DPO. In contrast, traditional RL frameworks typically rely on a fixed RM, making it difficult to update the RM in an online manner.
    \item \textbf{Multi-Round DPO Achieves RL-Level Performance with Lower Computational Cost}: Our final model DPO-VP, based on Qwen2.5-Math-7B, achieves comparable average performance to SimpleRL-Zero, PURE-VR and DPO-R1-Zero across 5 challenging benchmarks (48.2 vs 48.8/47.7/47.0), while requiring only a single 80GB GPU for training.
\end{enumerate}

\section{Methods}
\begin{figure}[t]
\centering
\includegraphics[width=1\textwidth]{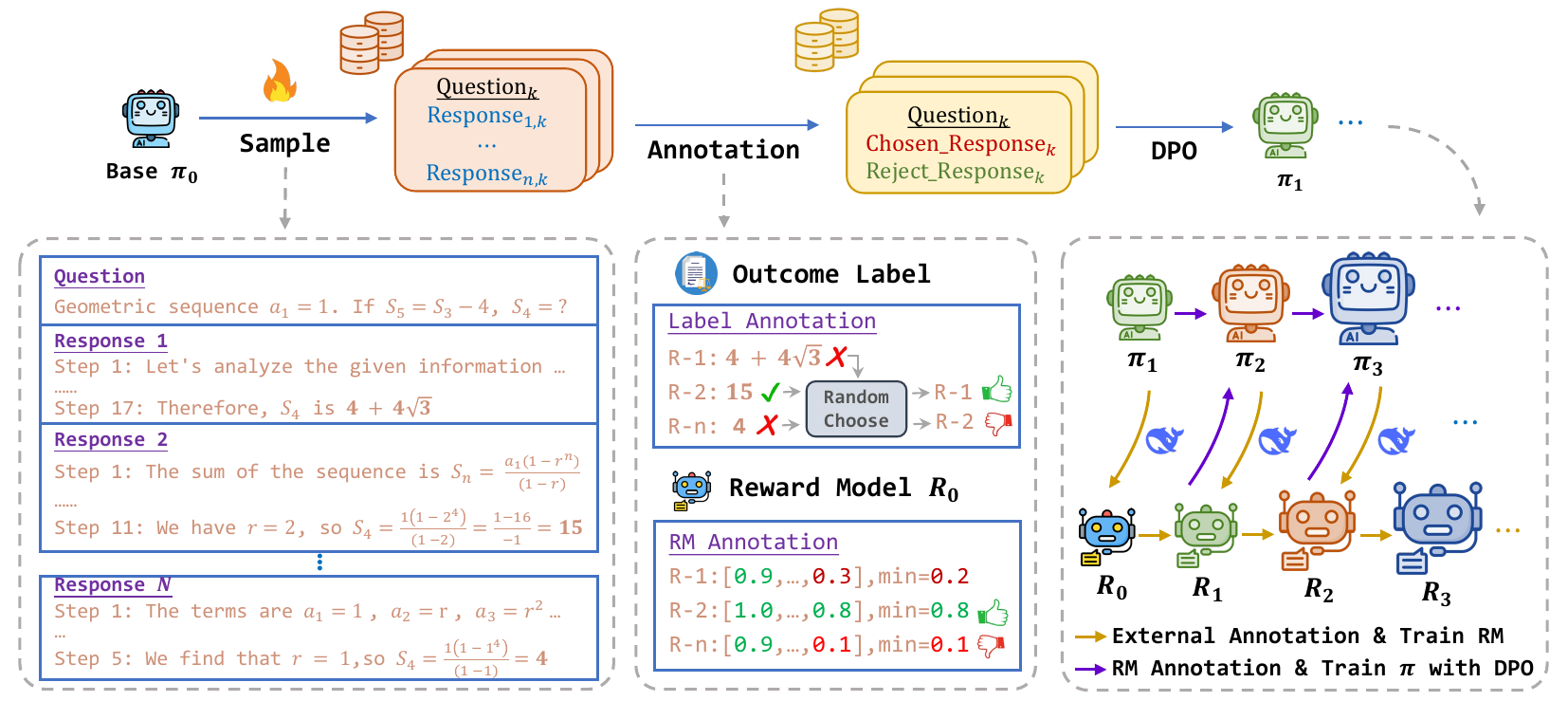}
\caption{Iterative DPO with Outcome Label and RM-Based Annotation}
\label{fig_main}
\end{figure}

The overall iterative DPO training framework is illustrated in Figure \ref{fig_main}.  
We first sample multiple responses from the base model, annotate them using rule-based or reward-based labeling. The policy is then optimized through DPO iterations, alternating between RM updates and policy refinement, enabling continuous self-improvement in reasoning ability.

\subsection{Iterative DPO with External Annotation}
For each question $Q_k$, the base model $\pi_0$ generates a set of candidate responses $\{ r_{1,k}, r_{2,k}, \dots, r_{n,k}\}$. 
Responses are evaluated using rule-based outcome labels or reward models (e.g. Process Reward Model, PRM or Outcome Reward Model, ORM) that assign preference scores.
Specifically, for ORM, we select the response with the highest score $s_{\text{ORM}}(r)$ as  $r^+$, and the response with the lowest score as $r^-$.
For PRM, where each step in a response is assigned a process reward score, we define the evaluation function $f$ as follows:
\begin{equation}
    f(s_{\text{PRM}}(r)) = \min \{s_{\text{PRM}}(r^1), \dots, s_{\text{PRM}}(r^n)\}
\end{equation}
where $n$ represents the total number of steps in response $r$. Then, the best response $r^+$ and the worst response $r^-$ are selected to construct the refined dataset, which is subsequently used to update the policy $\pi_0$ using DPO loss:
\begin{equation}
    \mathcal{L}_{\text{DPO}}(\pi) = \mathbb{E}_{(Q_k, r^+, r^-)} \left[ \log \sigma \left( \beta \cdot \left( \log \frac{\pi(r^+ \mid Q_k)}{\pi_0(r^+ \mid Q_k)} - \log \frac{\pi(r^- \mid Q_k)}{\pi_0(r^- \mid Q_k)} \right) \right) \right]
\end{equation}
where $\sigma$ is the sigmoid function and $\beta$ is a temperature scaling parameter.
The entire process is conducted over $N$ iterations, where the policy is progressively refined through repeated optimization. Finally, we obtain the final optimized policy $\pi_N$.

\subsection{Training Reward Model with Online Generated Data}
Furthermore, we can train the reward model using data generated by the online generator.
Taking PRM as an example, we leverage external annotators (e.g., DeepSeek-V3 \citep{liu2024deepseek}) to label the responses sampled from the current policy $\pi_t$. This process produces a new round of PRM training data, which is then used to fine-tune the previous PRM $R_{t-1}$, yielding an updated PRM $R_t$.
The training loss for updating the PRM is defined as follows:
\begin{equation}
\mathcal{L}_{\text{PRM}}(R_t) = \mathbb{E}_{(Q_k, r^+ , r^-)} \left[ -\log \sigma \left( R_t(Q_k, r^+) - R_t(Q_k, r^-) \right) \right]
\end{equation}

By iteratively updating the reward model and selecting the data for next epoch, we progressively improve its ability to assess process-level reasoning. 
In Section \ref{sec_e4}, we explore how this online training paradigm strengthens the PRM’s capability to capture and differentiate reasoning quality.

\section{Experiments}
\subsection{Setup}

In this section, we systematically investigate the impact of DPO on different base models. 
To ensure a comprehensive analysis, we examine models with varying parameter sizes and architectures, including Qwen2.5-3B \citep{yang2024qwen25}, Qwen2.5-7B \citep{yang2024qwen25}, LLaMA3.1 \citep{grattafiori2024llama}, and Qwen2.5-Math-7B \citep{yang2024qwen2}. 
The structure of the following sections is as follows:
\begin{itemize}[leftmargin=*]
\item In Section \ref{sec_e2}, we explore the simplest approach to improving base model performance by examining the effects of Chain-of-Thought (CoT) prompting and supervised fine-tuning (SFT) on different models, establishing a baseline for subsequent experiments.
\item In Section \ref{sec_e3}, we demonstrate that a single round of DPO with coarse filtering is sufficient to significantly enhance Qwen2.5’s mathematical reasoning performance through self-improvement.
\item In Section \ref{sec_e4}, we investigate how multi-round DPO and iterative reward model training enable continuous iterative improvement between the generator and the reward model.
\item In Section \ref{sec_e5}, we compare the final multi-round DPO results with state-of-the-art reinforcement learning methods and instruct models, analyzing potential differences in reasoning paradigms.
\end{itemize}

\subsection{CoT Prompts Improve Qwen2.5}
\label{sec_e2}

First, we need to ensure that the base model can align its output with a specified step-by-step format. 
To achieve this, we select problems from the training sets of GSM8K \citep{gsm8k} and MATH \citep{math} and reconstruct the reasoning paths in a structured step-by-step format using DeepSeek-V3 \citep{liu2024deepseek}. 
The specific prompt can be found in Figure \ref{a1_step_prompt} in the Appendix \ref{idx_prompt}.
This process facilitates the creation of a well-structured SFT dataset. 
Subsequently, we conduct two epochs of SFT on the base model.

\begin{wrapfigure}[14]{r}{0.4\textwidth}
    \centering
    \includegraphics[width=0.38\textwidth]{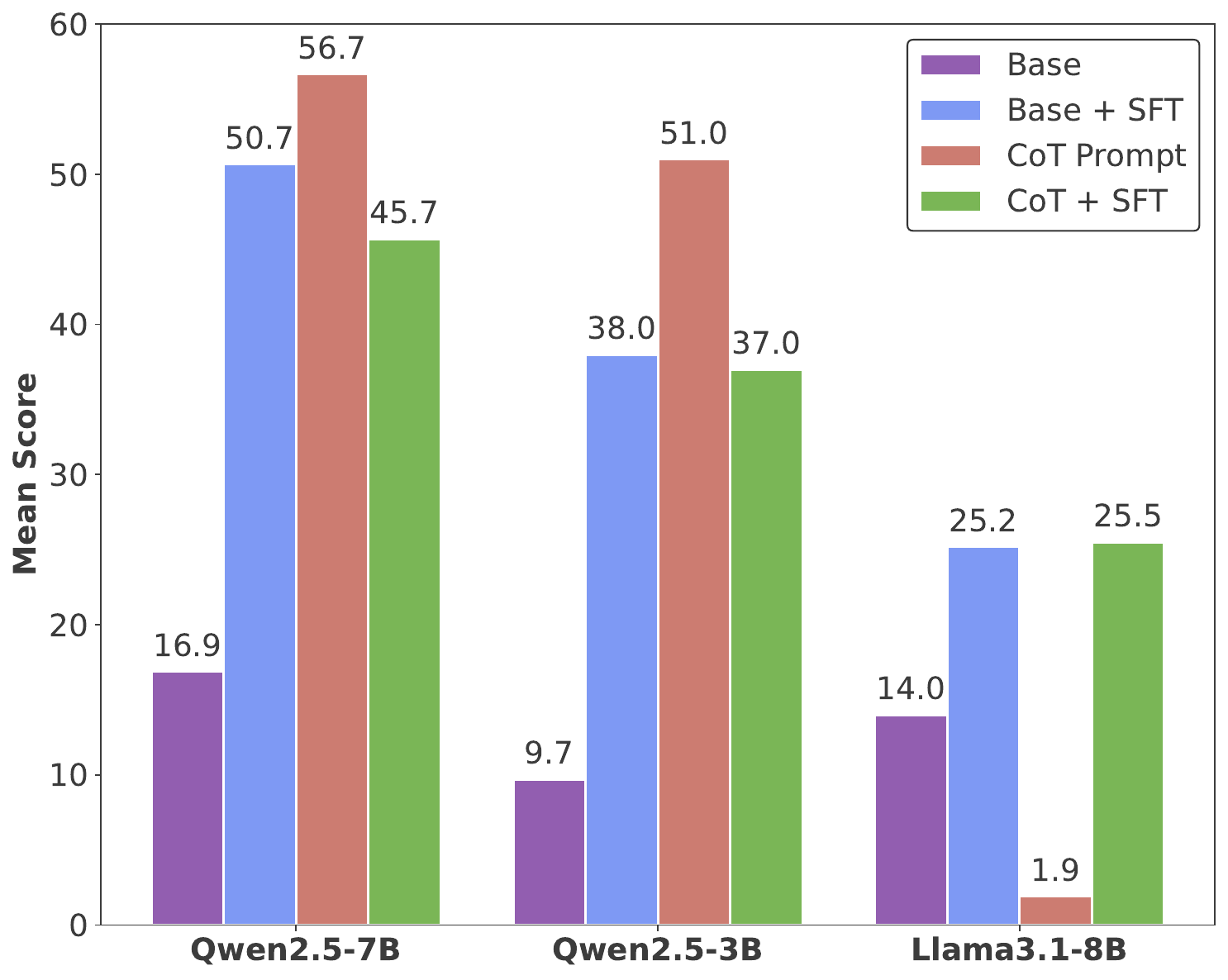}
    \caption{SFT \& CoT Comparison}
    \label{plot_sft}
\end{wrapfigure}

Meanwhile, inspired by \citet{kojima2022large}, we explore the use of a CoT prompt “\textit{Please reason step by step with steps separated by "\textbackslash n\textbackslash n", and put your final answer within \textbackslash boxed\{\}.}”, as a means to naturally guide the base model toward better adherence to the desired step-by-step format. 
The average greedy sampling scores across the 4 datasets (GSM8K, MATH500,Gaokao-2023-EN \citep{zhang2023evaluating} and  Minerva-Math \citep{yang2024qwen2}) are presented in Figure \ref{plot_sft}. 
"Base" denotes the approach where the model generates answers directly using the format “\textit{ Question: \{input\} \textbackslash n Answer: \{output\}.}” 
And "+SFT" indicates that the dataset was constructed using two different prompting strategies, followed by fine-tuning on the model.

The results indicate that for Qwen2.5, simply providing the CoT prompt is sufficient. 
Applying SFT, however, results in a reduction. 
This decline is probably attributed to the inferior quality of the GSM8K and MATH datasets.
In contrast, for LLaMA3.1, SFT is needed to ensure the model conforms to the designated format.
Therefore, \textbf{in the subsequent experiments, all training and evaluation processes are conducted using the CoT prompt, with LLaMA3.1 employing the CoT-SFT version.}

\subsection{Single-Round DPO with Coarse Filtering Improves Qwen2.5}
\label{sec_e3}
To explore the impact of DPO on enhancing base models, we first train an ORM and a PRM for each base model. 
We generate 4 responses per problem from the GSM8K and MATH training sets with Qwen2.5-7B, then evaluate each reasoning path with DeepSeek-V3 to obtain process-level reward labels, the evaluated prompts can be seen in Figure \ref{a2_reward_prompt} in the Appendix \ref{idx_prompt}.
The labels are filtered based on reference answers, retaining those that either match the reference with fully correct process annotations or differ from the reference with errors in process annotations. 
The filtered data are then used to fine-tune the base model, yielding the PRM.
Additionally, we construct preference data by pairing correct and incorrect responses for the same problem, then producing the ORM. 
Notably, our reward model relies on only tens of thousands of in-house data samples, making the training cost relatively low. 
Thus, it can serve as a coarse-grained filter for selecting self-improvement data.

After training the reward model, we resample 8 responses from the MATH training set using the base model. 
These responses are then filtered based on different rules to construct a DPO dataset with positive and negative sample pairs. 
Finally, we perform one round of DPO training on this dataset and evaluated various model variants on different datasets, including both in-distribution and out-of-distribution datasets. 
The results are recorded in Table \ref{tab_dpo_compare}. 
Specifically:
\begin{itemize}[leftmargin=*]
    \item \textbf{+Outcome Label}: Accuracy is determined by the outcome label, then a correct and an incorrect response are randomly selected as positive and negative samples, respectively.
    \item \textbf{+PRM \& +ORM}: The response with the highest RM score is selected as the positive sample, and the one with the lowest score as the negative sample. For PRM, the RM evaluation score is determined by the step with the lowest score.
    \item \textbf{+PRM with offset}: Additionally, high-PRM-score responses with incorrect answers, referred to as offset samples, are selected as negative samples to prevent the generator from exploiting PRM. We rank the responses by PRM scores in descending order, and among the top $k$ samples (where $k$ is the number of correct responses), incorrect responses are paired with the highest-scoring correct ones.
\end{itemize}

Additionally, we compare our results with closed-source models, including GPT-4o \citep{hurst2024gpt}, GPT-o1-mini \citep{jaech2024openai}, Claude-3.5-Sonnet \cite{2024Anthropic}, Qwen2.5-7B-Instruct \cite{yang2024qwen25}, and Qwen2.5-7B-DsV3-SFT, the latter being a version of Qwen2.5-7B SFT on GSM8K and MATH responses generated by DeepSeek-V3.

\begin{table}[thbp]
\centering
\renewcommand{\arraystretch}{1.2}
\setlength{\tabcolsep}{3pt}
\resizebox{\textwidth}{!}{
\begin{tabular}{l|l|c|ccc|ccc|ccc|ccc}
\toprule
\multirow{3}{*}{\textbf{Models}} & \multirow{3}{*}{\textbf{Methods}} & \multirow{3}{*}{\textbf{greedy-avg}} & \multicolumn{6}{c|}{\colorbox{lightyellow}{\textbf{In-Distribution Dataset}}} & \multicolumn{6}{c}{\colorbox{lightyellow}{\textbf{Out-of-Distribution Dataset}}} \\
& &  & \multicolumn{3}{c|}{\textbf{GSM8K}} & \multicolumn{3}{c|}{\textbf{MATH-500}} & \multicolumn{3}{c|}{\textbf{Gaokao-2023-EN}} & \multicolumn{3}{c}{\textbf{Minerva-Math}} \\
& &  & \textbf{greedy} & \textbf{mv@8} & \textbf{pass@8} & \textbf{greedy} & \textbf{mv@8} & \textbf{pass@8} & \textbf{greedy} & \textbf{mv@8} & \textbf{pass@8} & \textbf{greedy} & \textbf{mv@8} & \textbf{pass@8} \\
\midrule
GPT-4o & - & 72.5 & 96.4 & - & - & 76.6 & - & - & 71.4 & - & - & 45.6 & - & - \\
Claude-3.5-Sonnet & - & 72.2 & 96.4 & - & - & 71.1 & - & - & 72.2 & - & - & 48.9 & - & - \\
GPT-o1-mini & - & 76.7 & 96.8 & - & - & 90.0 & - & - & 70.9 & - & - & 48.9 & - & - \\
Qwen2.5-7B-Instruct & - & 68.7 & 92.4 & 91.4 & 97.8 & 76.4 & 73.4 & 88.2 & 64.2 & 61.0 & 78.7 & 41.9 & 20.2 & 41.2 \\
Qwen2.5-7B-DsV3-SFT & - & 62.5 & 89.5 & 92.0 & 96.7 & 72.6 & 78.0 & 87.4 & 60.0 & 67.0 & 76.1 & 27.9 & 29.4 & 51.4 \\
\midrule
\multirow{5}{*}{Qwen2.5-7B} & Base & 56.7 & 86.4 & 92.1 & 97.2 & 64.2 & 74.6 & 87.8 & 55.6 & 61.3 & 76.6 & 20.6 & 18.4 & 35.3 \\
 & +Outcome Label & 63.0 & 89.8 & 93.5 & 97.3 & 74.2 & 78.2 & 87.6 & 62.9 & \textbf{67.3} & 76.9 & 25.0 & 24.6 & 41.2 \\
 & +ORM & 62.2 & 89.5 & 93.4 & 97.2 & 73.6 & \textbf{79.6} & 88.8 & 61.6 & 66.2 & 78.2 & 23.9 & \textbf{26.8} & \textbf{47.7} \\
 & +PRM & 62.8 & \textbf{90.5} & 93.3 & 96.9 & 74.6 & 78.4 & 88.6 & 61.0 & 66.5 & 78.4 & 25.0 & 20.2 & 37.5 \\
 & +PRM with offset & \textbf{64.0} & 90.3 & \textbf{93.9} & \textbf{97.5} & \textbf{75.8} & 78.6 & \textbf{88.8} & \textbf{63.9} & 65.7 & \textbf{78.7} & \textbf{25.9} & 23.9 & 37.9 \\
    &  \cellcolor{lightblue}Best Improve & \cellcolor{lightblue}+7.3 & \cellcolor{lightblue}+4.1 & \cellcolor{lightblue}+1.8 & \cellcolor{lightblue}+0.3 & \cellcolor{lightblue}+11.6 & \cellcolor{lightblue}+5.0 & \cellcolor{lightblue}+1.0 & \cellcolor{lightblue}+8.3 & \cellcolor{lightblue}+6.0 & \cellcolor{lightblue}+2.1 & \cellcolor{lightblue}+5.3 & \cellcolor{lightblue}+8.4 & \cellcolor{lightblue}+12.4 \\
\midrule
\multirow{5}{*}{Qwen2.5-3B} & Base & 49.0 & 74.7 & 86.9 & 95.6 & 61.2 & 68.2 & 81.4 & 42.1 & 56.6 & 71.2 & \textbf{18.0} & \textbf{18.8} & 31.6 \\
 & +Outcome Label & 53.7 & 82.5 & 87.5 & 94.6 & 63.8 & \textbf{70.4} & 82.8 & 52.5 & 55.6 & 70.6 & 15.8 & 16.9 & 29.7 \\
 & +ORM & \textbf{54.1} & 83.8 & \textbf{89.0} & 95.4 & \textbf{64.4} & 69.0 & \textbf{83.6} & \textbf{54.8} & 56.6 & \textbf{73.3} & 13.2 & 17.3 & \textbf{32.7} \\
  & +PRM & 53.9 & 83.2 & 88.4 & 95.4 & \textbf{64.4} & \textbf{70.4} & 83.0 & 53.0 & \textbf{58.2} & 71.2 & 15.1 & 16.2 & \textbf{32.7} \\
  & +PRM with offset & 53.9 & \textbf{83.9} & 88.2 & \textbf{95.9} & 63.2 & 68.4 & 82.0 & 53.2 & 57.9 & 71.2 & 15.4 & 16.2 & 30.5 \\
  &  \cellcolor{lightblue}Best Improve & \cellcolor{lightblue}+5.1 & \cellcolor{lightblue}+9.2 & \cellcolor{lightblue}+2.1 & \cellcolor{lightblue}+0.3 & \cellcolor{lightblue}+3.2 & \cellcolor{lightblue}+2.2 & \cellcolor{lightblue}+2.2 & \cellcolor{lightblue}+12.7 & \cellcolor{lightblue}+1.6 & \cellcolor{lightblue}+2.1 & \cellcolor{lightblue}-2.2 & \cellcolor{lightblue}-1.5 & \cellcolor{lightblue}+1.1 \\
\midrule
\multirow{5}{*}{LLaMA3.1-8B} & Base & 25.4 & 47.4 & 58.4 & 81.8 & \textbf{21.6} & 27.6 & 48.2 & 19.5 & \textbf{26.7} & 46.2 & 13.6 & 14.7 & \textbf{32.0} \\
 & +Outcome Label & 25.7 & 48.6 & 61.9 & 81.8 & 20.0 & 27.4 & 48.4 & 19.2 & 25.7 & 45.7 & \textbf{15.1} & \textbf{15.8} & 28.7 \\
 & +ORM & 25.4 & 48.7 & 60.3 & 83.1 & 20.8 & \textbf{28.8} & 50.4 & 19.0 & 24.7 & 42.1 & 12.9 & 11.8 & 31.6 \\
 & +PRM & 26.4 & 48.2 & 58.9 & 79.1 & 20.4 & 24.2 & 47.8 & \textbf{22.1} & 26.0 & \textbf{48.2} & 14.7 & 10.0 & 25.7 \\
 & +PRM with offset & \textbf{26.5} & \textbf{49.8} & \textbf{62.5} & \textbf{84.3} & \textbf{21.6} & 28.6 & \textbf{51.8} & 21.6 & 24.9 & 47.5 & 12.9 & 10.3 & 31.3 \\
  &  \cellcolor{lightblue}Best Improve & \cellcolor{lightblue}+1.1 & \cellcolor{lightblue}+2.4 & \cellcolor{lightblue}+4.1 & \cellcolor{lightblue}+2.5 & \cellcolor{lightblue}+0.0 & \cellcolor{lightblue}+1.2 & \cellcolor{lightblue}+3.6 & \cellcolor{lightblue}+2.6 & \cellcolor{lightblue}-0.7 & \cellcolor{lightblue}+2.0 & \cellcolor{lightblue}+1.5 & \cellcolor{lightblue}+1.1 & \cellcolor{lightblue}-0.4 \\
\bottomrule
\end{tabular}
}
\caption{Performance Comparison Across Different Models and DPO Data Filtering Methods. \textbf{Bold} values indicate the best results, and highlighted rows represent the best improvements.}
\label{tab_dpo_compare}
\end{table}

Based on the results in Table \ref{tab_dpo_compare}, we derive the following findings:  

\begin{enumerate}[leftmargin=*]
\item \textbf{DPO can rapidly enhance performance, with improvement strongly correlated to the base model's capability.} For Qwen2.5-7B, a single round of DPO using self-generated data brings the best variant, surpassing Qwen2.5-7B-DsV3-SFT and approaching Qwen2.5-7B-Instruct, which has been fine-tuned on a large dataset. However, for LLaMA3.1-8B, DPO has little effect on self-improvement. 
\item \textbf{Even coarse-grained filtering significantly boosts performance. Leveraging a reward model for supervision and incorporating additional offset negative samples may improve training efficiency.} For Qwen2.5-7B, PRM with offset achieves the highest performance without requiring extra sampling.
\end{enumerate}

\subsection{Iterative Improvement of Generator and RM can be achieved via Multi-Round DPO}
\label{sec_e4}
An interesting question is: can the generator and RM achieve mutual improvement through multiple rounds of DPO? The answer is certainly  YES!

Starting with Qwen2.5-7B-base, we greedily sample responses from the MATH training set and annotate using DeepSeek-V3 to initialize PRM-base. 
Then, we apply the "DPO+PRM" filtering method from the previous section to select data for training the current generator via DPO. 
After updating the generator, we continue with online greedy sampling on MATH, annotation, and PRM updates. 
Just as illustrated in the bottom right of Figure \ref{fig_main}, this iterative process allows the generator and RM to evolve together over three epochs.

\begin{figure*}[htbp]
\centering
\subfloat[Generator Score.]{
\includegraphics[width=.33\linewidth]{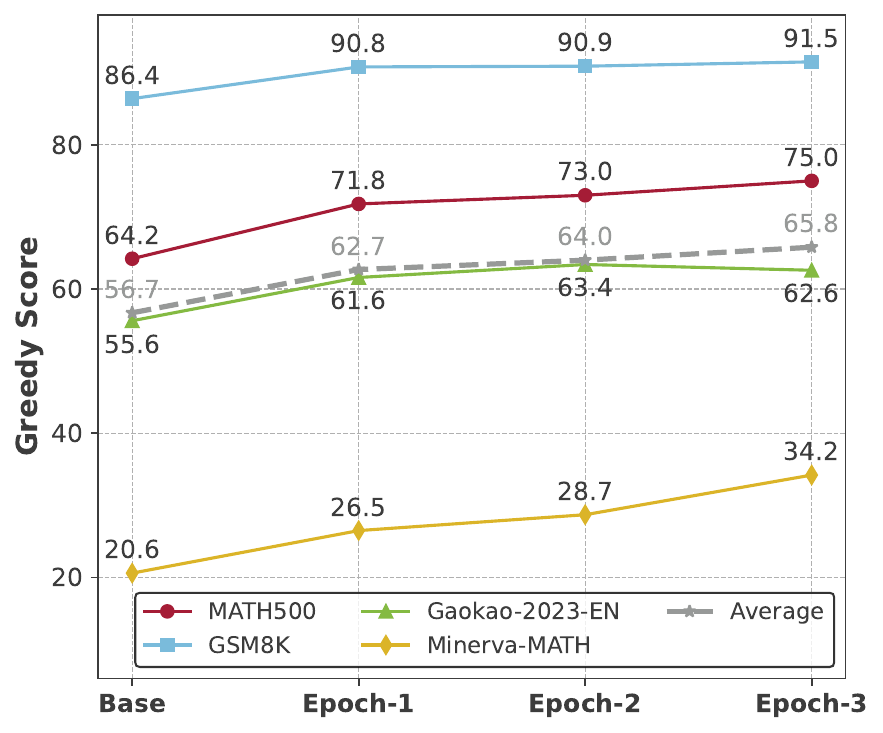}
\label{2_1_plot_self_improve_generator}
}
\subfloat[PRM Evaluation.]{
\includegraphics[width=.33\linewidth]{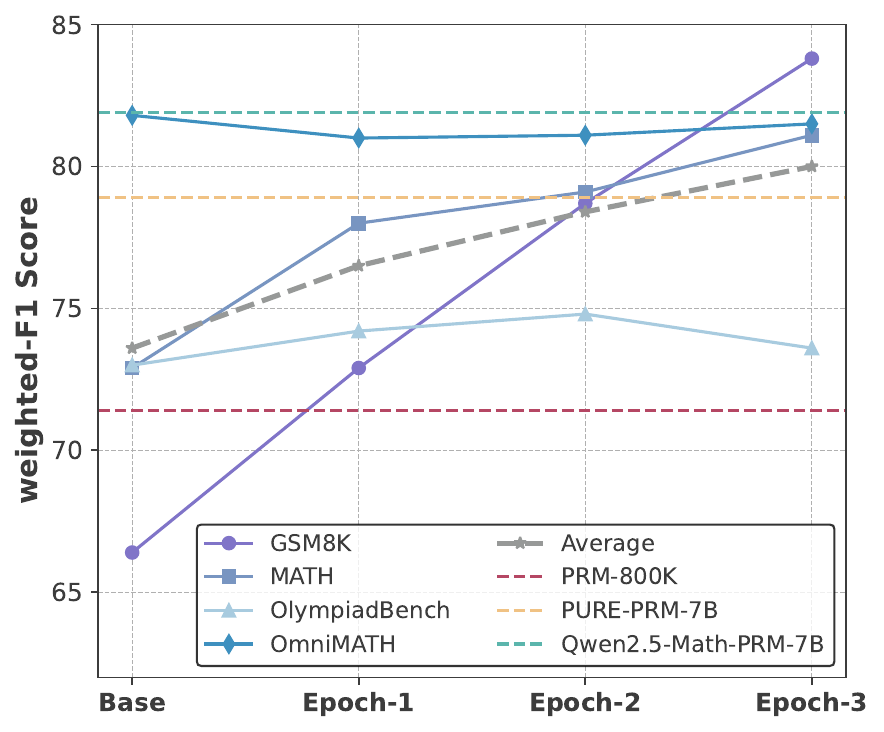}
\label{2_2_plot_self_improve_prm}
}
\subfloat[Test-time Scaling.]{
\includegraphics[width=.33\linewidth]{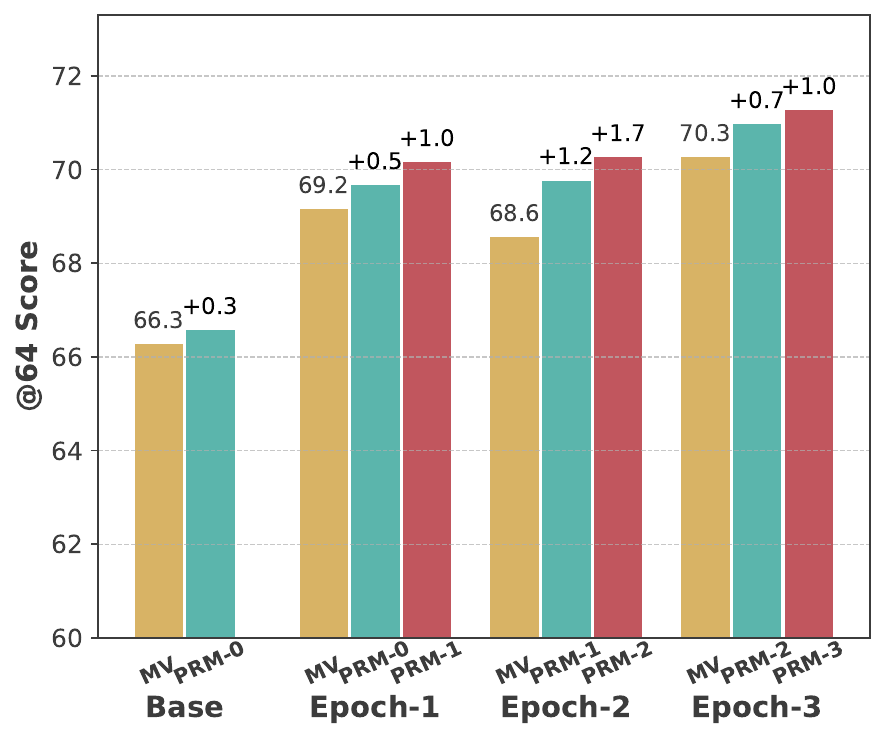}
\label{2_3_plot_self_improve_iter}
}
\caption{Performance Evolution of Generator and PRM Across Multiple DPO Epochs.}
\label{fig:motivation}
\end{figure*}

We conduct a detailed evaluation of the iteratively trained generator and PRM:

\begin{enumerate}[leftmargin=*]
\item \textbf{Generator Score}: As shown in Figure \ref{2_1_plot_self_improve_generator}, we observe continuous performance improvement across different datasets. \textbf{The average evaluation score increases from 62.7 in the first epoch to 65.8 in the third epoch, surpassing the best average score reported in Table \ref{tab_dpo_compare}.}
\item \textbf{PRM Evaluation}: We assess the ability of PRM  to identify incorrect steps using ProcessBench \citep{zheng2024processbench}. Given the imbalance between positive and negative samples in ProcessBench, we compute the weighted-F1 score \citep{hinojosa2024performance} to reflect error detection performance. In addition to self-comparison, we benchmark against several baselines: 
\begin{enumerate}[leftmargin=*, label=(\alph*)]
\item\underline{PRM-800K}: A PRM fine-tuned on PRM800K \citep{lightman2023let} dataset with Qwen2.5-7B.  
\item\underline{PURE-PRM-7B}: A PRM derived from \citet{cheng2025pure}, trained with PRM800K through a two-stage MLP and full fine-tuning. 
\item\underline{Qwen2.5-Math-PRM-7B}: The SOTA PRM officially released by Qwen \citep{zheng2024processbench}, trained on a large in-house dataset. 
\end{enumerate}
The results show that \textbf{our PRM surpasses PRM-800K and, by the third epoch, outperforms PURE-PRM-7B, while slightly below Qwen2.5-Math-PRM-7B.} \textbf{And demonstrate that PRM can continuously improve through iterative training with the generator at a low cost — each epoch requires only 7.5K DeepSeek-V3 queries.} Detailed experimental settings can be found in Appendix \ref{idx_processbench}.
\item \textbf{PRM in Test-time Scaling}: We further analyze the role of PRM in test-time scaling. As shown in Figure \ref{2_3_plot_self_improve_iter}, we compare majority vote (MV) \citep{wangself2023} and PRM@64. We find that PRM@64 outperforms MV in every epoch, and \textbf{the PRM trained with online sampled data from the current generator further improves over the previous epoch’s PRM@64 performance}. A detailed performance table is provided in Appendix \ref{idx_tts}.
\end{enumerate}

\subsection{Multi-Round DPO with Verifiable Pairs Achieves RL-Level Performance Efficiently}
\label{sec_e5}
\subsubsection{Setup and Main Results}
Finally, we conduct multi-round DPO on the advanced Qwen2.5-Math-7B, aiming to benchmark against RL-based approaches under the same base model and training data. 
Inspired by DeepSeek-R1 \citep{zeng2025simplerl}, we observe that the verifiable reward (VR) is a 0/1 discrete variable, indicating whether the output meets predefined correctness criteria. 
This aligns with the original DPO framework, where correct samples are classified as positive and incorrect ones as negative, forming verifiable pairs (VP) for training.
By applying VP-based filtering and gradual temperature scaling (inspired by \citet{xi2024enhancing}), we conduct 6 epochs of DPO. 
The final model, \underline{Qwen2.5-7B-DPO-VP}, is compared against several state-of-the-art RL-based methods:
\begin{enumerate}[leftmargin=*, label=(\alph*)]
\item\underline{Qwen2.5-Math-7B-Instruct} \citep{yang2024qwen2}: An Instruct model by Qwen with strong mathematical reasoning capabilities.
\item\underline{rStar-Math-7B} \citep{guan2025rstar}: A model trained with a large amount of self-evolution MCTS data and filtered with process preference model (PPM).  
\item\underline{Eurus-2-7B-PRIME} \citep{cui2025process}: A model trained by process RL with implicit process rewards. 
\item\underline{Qwen2.5-7B-GRPO-Zero} \citep{shao2024deepseekmath}:  A model trained by GRPO with verifiable rewards without any extra SFT data.
\item\underline{Qwen2.5-7B-Simple-RL-Zero/Zoo} \citep{zeng2025simplerlzoo,zeng2025simplerl}:  A model trained by PPO with verifiable rewards without any extra SFT data.  Zero and Zoo refer to two versions released at different time points.
\item\underline{Qwen2.5-7B-PURE-VR} \citep{cheng2025pure}:  A model trained by RL with verifiable rewards and min-form credit assignment without any extra SFT data.
\item\underline{Qwen2.5-7B-DPO-R1-Zero} \citep{zhang2025dpor1}: A concurrent work that applied iterative DPO without any extra SFT data.
\end{enumerate}

To align with existing RL methods, following \citet{zeng2025simplerl} and \citet{cheng2025pure}, we select 8K Level 3-5 questions from the MATH as self-improvement data, and additionally evaluate on Olympiad-level benchmarks, including AMC23, AIME24, and OlympiadBench \cite{he2024olympiadbench}.
Furthermore, we conduct additional comparisons on 5 out-of-domain datasets to assess generalization performance.
In addition, we compare with Qwen2.5-7B-DPO-R1-Zero \citep{zhang2025dpor1} that applies iterative DPO on 200K questions (more than 8K questions) from the NuminaMath dataset \citep{li2024numinamath}. 

Moreover, we employ annealed sampling to enhance response diversity. Specifically, the sampling temperature $t$ is set to 0.7 for the first three epochs, increased to 1.0 during epochs 4–5, and further raised to 1.2 in the final epoch. 
This annealing schedule is motivated by the observation that model performance nearly converges after three epochs, and increasing the temperature in later stages allows for fine-tuning on more diverse samples, leading to additional gains.

Table \ref{tab_pass1-acc} presents the pass@1 accuracy of different models, while Table \ref{tab_data_gpu_comparison} further analyzes training data consumption and GPU usage. 

\begin{table}[htbp]
\centering
\resizebox{\textwidth}{!}{
\begin{tabular}{lcccccc}
    \toprule
    \textbf{pass@1 acc} & \textbf{MATH500} & \textbf{Minerva-Math} & \textbf{Olympaidbench} & \textbf{AMC23} & \textbf{AIME24} & \textbf{Avg.} \\
    \midrule
    Qwen2.5-Math-7B $^*$ & 64.8 & 15.4 & 25.6 & 37.5 & 16.7 & 32.0 \\
    Qwen2.5-Math-7B-Instruct $^*$ & \textbf{83.2} & 33.5 & 38.4 & \underline{62.5} & 20.0 & 47.5 \\
    rStar-Math-7B $^\wedge$ & 78.4 & - & \textbf{47.1} & 47.5 & \textbf{26.7} & - \\
    Eurus-2-7B-PRIME $^*$ & 74.0 & \textbf{39.7} & 35.6 & 57.5 & 23.3 & 46.0 \\
    \textbf{Qwen2.5-7B-GRPO-Zero} $^\wedge$ & 76.2 & 32.7 & 38.1 & 55.0 & 16.7 & 43.7 \\
    \textbf{Qwen2.5-7B-Simple-RL-Zero} $^*$ & 78.0 & 33.1 & 36.6 & 60.0 & \textbf{26.7} & 46.9 \\
    \textbf{Qwen2.5-7B-Simple-RL-Zoo} $^*$ & \underline{80.4} & \textbf{39.7} & 38.8 & 57.5 & \textbf{26.7} & \textbf{48.7} \\
    \textbf{Qwen2.5-7B-PURE-VR} $^*$ & \underline{79.8} & \underline{36.8} & \underline{41.9} & 60.0 & 20.0 & 47.7 \\
    \textbf{Qwen2.5-7B-DPO-R1-Zero} $^\wedge$ & 76.8 & 30.9 & 37.9 & \underline{62.5} & \textbf{26.7} & 47.0 \\
    \cellcolor{lightblue}\textbf{Qwen2.5-7B-DPO-VP} & \cellcolor{lightblue}74.8 & \cellcolor{lightblue}35.3 & \cellcolor{lightblue}36.9 & \cellcolor{lightblue}\textbf{67.5} & \cellcolor{lightblue}\textbf{26.7} & \cellcolor{lightblue}\underline{48.2} \\
    \bottomrule
\end{tabular}
}
\caption{Pass@1 Accuracy Results Across Different Benchmarks. \textbf{Bold} values indicate the best results, and \underline{underline} value indicate the second one. All models are fine-tuned based on the Qwen2.5-Math-7B. \textbf{Bold} models represent those that were adjusted using the self-improvement method with exactly the same prompts. The results with $*$ are from our reproduced, and the results with $\wedge$ are derived from the corresponding technical report.}
\label{tab_pass1-acc}
\end{table}

\begin{table}[h]
    \centering
    \resizebox{\textwidth}{!}{
    \renewcommand{\arraystretch}{1.2}
    \rowcolors{2}{gray!15}{white} 
    \begin{tabular}{p{3cm} p{3cm} p{3cm} p{3cm} p{3.2cm} p{3cm} |p{3cm}}
        \toprule
        & \textbf{Qwen2.5-Math-7B-Instruct} & \textbf{rStar-Math-7B} & \textbf{Eurus-2-7B-PRIME} & \textbf{Qwen2.5-7B-SimpleRL-Zero} & \textbf{Qwen2.5-7B-PURE-VR} & \textbf{Qwen2.5-7B-DPO-VP} \\
        \midrule
        \textbf{Base Model} & Qwen2.5-Math-7B & Qwen2.5-Math-7B & Qwen2.5-Math-7B & Qwen2.5-Math-7B & Qwen2.5-Math-7B & Qwen2.5-Math-7B \\
        \textbf{SFT Data} & 2.5M (open-source and in-house) & $\sim$7.3M (MATH, NuminaMath, etc.) & 230K & 0 & 0 & 0 \\
        \textbf{RM Data} & 618K (in-house) & $\sim$7K (in-house) & 0 & 0 & 0 & 0 \\
        \textbf{RM} & Qwen2.5-Math-RM (72B) & None & Eurus-2-7B-SFT & None & None & None \\
        \textbf{Self-improve Method} & RL + ORM & MCTS + PPM & RL + PRM & RL + VR & RL + VR & DPO + VR \\
        \textbf{Self-improve Data} & 66K & $\sim$3.647M & 150K & \textbf{8K MATH} & \textbf{8K MATH} & \textbf{8K MATH} \\
        \textbf{GPUs \& Training Times} & - & 80 H100 for few days (at most) & 8 A100 for few days & 32 H100 for 1.5 days & 8 A100 for 1 day& \textbf{4 A800 for $<$1 day or 1 A800 for 2 days} \\
        \bottomrule
    \end{tabular}
    }
\caption{Data and GPUs Comparison of Different Models.}
\label{tab_data_gpu_comparison}
\end{table}

Table \ref{tab_pass1-acc} and \ref{tab_data_gpu_comparison}, we derive the following conclusions:  
\begin{enumerate}[leftmargin=*]
\item \textbf{Multi-round DPO can achieve mathematical reasoning capabilities comparable to those of RL-based methods.} The final model attains an average score of 48.2 across five mathematical reasoning benchmarks, which is similar to the performance of Qwen2.5-Math-7B-Instruct and other RL-based approaches with the some training data conditions. Compared to Qwen2.5-7B-DPO-R1-Zero, our model still achieves a higher average score. We hypothesize that this may be attributed to the increased response diversity introduced by the annealed sampling strategy.
\item \textbf{Multi-round DPO requires significantly fewer computational resource can even be executed on a single GPU.} Our sampling and training are conducted on four A800 GPUs using data parallelism. The full pipeline can be run on a single 80GB GPU or lower. In our setup with 4 A800 GPUs, each sampling round took 2–2.5 hours with vLLM \citep{kwon2023efficient}, training took 1 hour per round, and the full process completed in about 1 day. On a single GPU, the full process would take around 3 days.
\end{enumerate}

\subsubsection{More Analysis}

\paragraph{Token Length Dynamics and Accuracy Trends}

\begin{figure}[htbp]
\centering
\includegraphics[width=.9\textwidth]{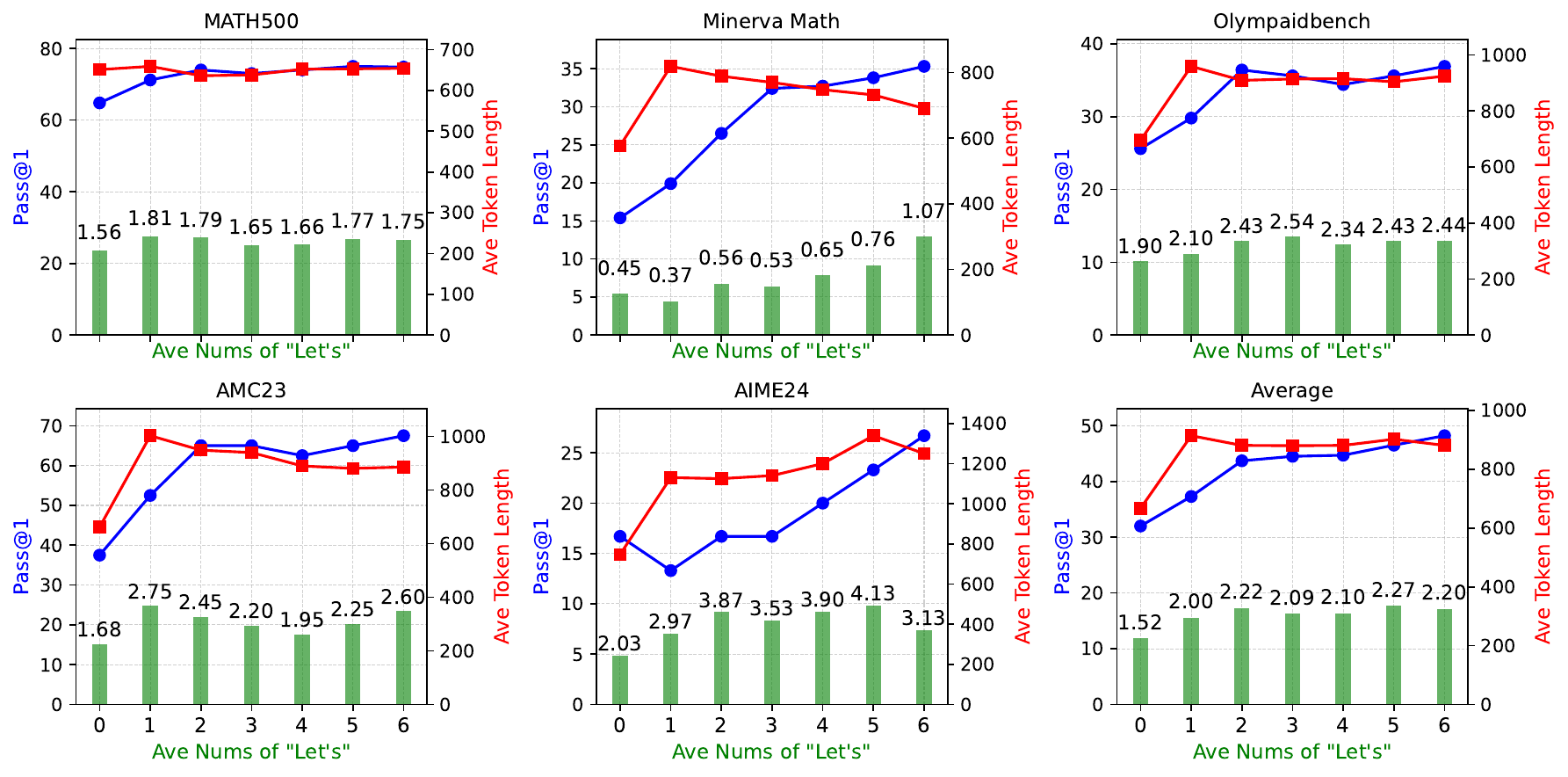}
\caption{Pass@1 Accuracy, Token Length, and Self-Evaluation Trends Across Datasets.}
\label{plot_acc_length}
\end{figure}

First, we analyze token length variations throughout training. In Figure \ref{plot_acc_length}, we illustrate the accuracy progression of the iterative process across different datasets in relation to the average token length. 
The results indicate a steady improvement in accuracy, while the \textbf{token length during inference did not exhibit an initial increase followed by a decline. Instead, it remains relatively stable within a consistent range.}
When comparing different models, we observe that their output lengths are generally similar. 
Detailed results are provided in Table \ref{tab_avg_token_length} in Appendix \ref{idx_dpovp_ana}.

\paragraph{Self-Reflection and Reasoning Paradigms}
We do not observe a significant "Aha Moment," which aligns with the findings of PURE \citep{cheng2025pure}. 
Even when using CoT prompts, Qwen2.5-Math-7B frequently relies on Python-based solutions, despite not actually executing the code.
Additionally, we identify a recurring self-evaluation pattern, such as \textit{”Let's re-evaluate ..."}, which is already present in the base model. 
This suggests that DPO does not inherently introduce self-reflection capabilities but instead strengthens existing tendencies through reward optimization, as shown in Figure \ref{plot_acc_length}.
When comparing several different models, Table \ref{tab_lets_occurrence} in Appendix \ref{idx_dpovp_ana} shows that Qwen2.5-Math-7B-Instruct exhibits significantly fewer occurrences of \textit{"Let's"}, indicating a shift in reasoning paradigms.

\subsection{Ablation Study on the DPO Sampling Noisy}

\begin{wrapfigure}[16]{r}{0.35\textwidth}
    \centering
    \includegraphics[width=0.33\textwidth]{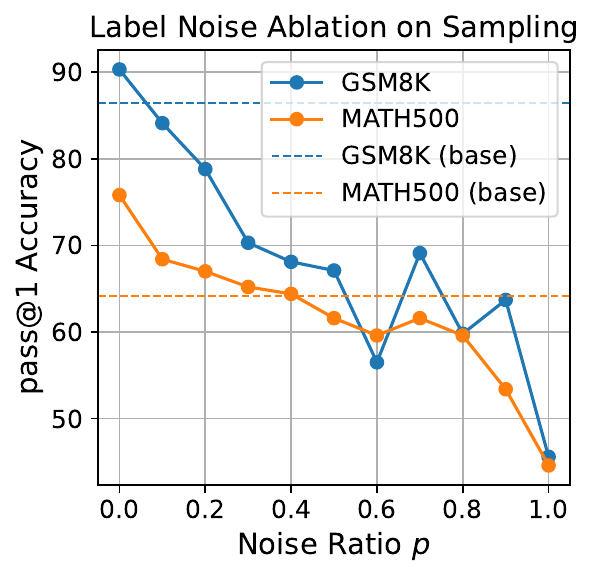}
    \caption{Label Noise Ablation}
    \label{plot_label_noisy}
\end{wrapfigure}

To evaluate the robustness of our DPO framework under noisy supervision, we conduct a controlled label noise ablation. Specifically, we simulate noise in the preference data used to train the Preference Reward Model (PRM) under the \textit{PRM with offset} setting. Following \citet{tu2025dataset}, for each positive–negative response pair, we apply a \textbf{random flipping probability} $p \in [0, 1]$ that reverses the preference direction with probability $p$, simulating increasing levels of annotation noise. For example, $p=0.5$ corresponds to fully random pairwise supervision, while $p=1.0$ denotes complete inversion of the labels.

We test the impact of this noise injection on the final model performance, measured by pass@1 accuracy after one round of DPO on GSM8K and MATH500 using Qwen2.5-7B. The results are shown in Figure \ref{plot_label_noisy}.
These results clearly demonstrate that model performance degrades as the preference supervision becomes less accurate. While DPO appears tolerant to small levels of label noise, performance drops significantly under higher noise levels. This provides empirical evidence that maintaining reasonably accurate verifier signals is critical for effective preference optimization.

\paragraph{Generalization}
We conduct evaluations on 5 additional Out-of-Domain datasets: 
\begin{enumerate}[leftmargin=*, label=(\alph*)]
\item \underline{CMATH}: A Chinese mathematics dataset.
\item \underline{MMLU\_STEM}: A subset of MMLU focusing on science subjects like physics, biology.
\item \underline{HumanEval}: A benchmark evaluating code generation through Python tasks.
\item \underline{LiveCodeBench}: A coding benchmark for real-time code editing and debugging.
\item \underline{RACE}: A reading benchmark from English exams testing reasoning.
\end{enumerate}

The first two datasets are evaluated using the Qwen-Math codebase \citep{yang2024qwen2}, with evaluation parameters consistent with those used in Appendix~\ref{idx_dpovp}. The remaining three datasets are evaluated via EvalScope\footnote{\url{https://github.com/modelscope/evalscope}} using its default configuration settings. The results are summarized in Table~\ref{tab_ood_dataset}.

\begin{table}[htbp]
    \centering
    \resizebox{1\textwidth}{!}{
    \begin{tabular}{lccccc|c}
        \toprule
        \textbf{pass@1 acc} & \textbf{CMATH} & \textbf{MMLU\_STEM} & \textbf{HumanEval} & \textbf{Live\_Code\_Bnech} & \textbf{RACE} & \textbf{Average}\\
        \midrule
        Qwen2.5-Math-7B & 64.0 & 58.3 & 40.9 & 5.1 & 61.6 & 46.0\\
        Eurus-2-7B-PRIME & 72.5 & 42.8 & 41.5& 21.2& 62.6 & 48.1\\
        Qwen2.5-7B-Simple-RL-Zero & 70.3 & 59.1 & 51.8 & 21.4 & 61.1 & 52.7 \\
        Qwen2.5-7B-Simple-RL-Zoo & 89.0 & 35.9 & 53.1 & 17.9 & 63.9 & 52.0 \\
        Qwen2.5-7B-PURE-VR & 69.5 & 58.8 & 43.3 & 6.4 & 61.3 & 47.9 \\
        \rowcolor{lightblue} \textbf{Qwen2.5-7B-DPO-VP} & 68.5 & 59.3 &50.6 & 17.3 & 61.2 & 51.4\\
        \bottomrule
    \end{tabular}
    }
    \caption{Performance comparison on Out-of-Domain datasets.}
    \label{tab_ood_dataset}
\end{table}

Nearly all methods lead to slight performance improvements on the out-of-domain (OOD) benchmarks, indicating enhanced generalization capabilities. Our proposed DPO-VP notably increases the average OOD score from 46.0 to 51.4, achieving gains comparable to those of RL-based methods. This suggests that DPO fine-tuning on mathematical tasks can also effectively enhance generalization to other domains.

\section{Related Works}

\paragraph{Self-improvement of LLMs} 
Self-improvement primarily occurs during the training phase, where the model enhances its own performance by evaluating and refining its outputs without relying on external human supervision \citep{xi2024enhancing,kumar2025llm}. 
It is generally believed that LLMs are incapable of achieving self-improvement without any external information \citep{huanglarge2024, tyen2024llms}.
Consequently, an increasing number of methods leverage external supervision signals help self-improvement, for example, outcome labels for mathematical or programming problems \citep{mcaleese2024llm, gao2024llm}, process reward models \citep{zhang2024generative,fu2025rlae}, value function \citep{wang2024q} or LLM-as-a-judge \citep{liself2024,zhang2025process} frameworks.

\paragraph{Improving Reasoning with DPO} 
Recent works have significantly extended DPO-based reasoning \citep{xu2024dpo,zhang2025dpor1}. Iterative DPO with rigorous theoretical analysis was proposed by \citet{xiong2024iterative}, and further generalized to multi-turn and tool-integrated scenarios \citep{xiong2024building}. Flow-DPO \citep{deng2024flow} introduced online multi-agent learning for mathematical reasoning, while iterative length-regularized DPO \citep{liu2024iterative} enables 7B models to approach GPT-4-level reasoning.
Self-play and self-training pipelines have also been explored, demonstrating that DPO and preference-based methods can improve LLMs beyond human-labeled data \citep{chen2024self, singh2023beyond}. Additionally, implicit reward bootstrapping with DPO has shown promise for further alignment without explicit reward models \citep{chen2024bootstrapping}. Aya Expanse \citep{dang2024aya} illustrates how such alignment advances benefit multilingual and cross-cultural reasoning.
In contrast, our work focuses on data selection strategies for DPO and proposes an iterative training framework for generator-verifier models, enabling mutual enhancement between generation and reward modeling.

\paragraph{Constructing Efficient Reasoning Models} 
In early 2025, with the release of DeepSeek-R1 \citep{guo2025deepseek}, efficient post-training paradigms gained significant attention \citep{trung2024reft,tu2025learning}.  
SimpleRL-Zero \citep{zeng2025simplerl} achieved remarkable  performance through RL, training on just 8K challenging mathematical problems with rule-based rewards.
\citet{xiong2025self} integrated reflective vocabulary into generative RM to enable selective RL-based correction.  
PURE \citep{cheng2025pure} revisited stepwise reward credit assignment in RL.  
Logic-RL \citep{xie2025logic} refined the penalty function for formatting errors, leading to improved logical reasoning capabilities.  
From the data utilization perspective, LIMO \citep{ye2025limo} and S1 \citep{muennighoff2025s1} demonstrated that fine-tuning with carefully curated long-chain reflective data can effectively amplify the latent reasoning capabilities.
From the reasoning paradigm perspective, CFT \citep{wang2025critique} encouraged models to develop critical thinking during SFT.
Our work focuses on achieving RL-based reasoning with significantly fewer computational resources. We demonstrate that, with a strong base model, coarse-grained selection is enough for performance enhancement, achieving competitive results with reduced computational overhead. 
Our overall training framework is more resource-efficient than the widely adopted SFT+RL pipeline, while outperforming concurrent iterative DPO methods.

\section{Conclusion}
In this paper, we empirically study DPO for enhancing LLM reasoning, showing that single-round DPO with coarse filtering improves performance, while multi-round DPO enables iterative enhancement of both the generator and RM. 
Our results demonstrate that multi-round DPO achieves RL-level performance with significantly lower computational costs, requiring only a single 80GB GPU. 
However, in large-scale deployment and long-term training, DPO lacks exploration and sustained improvement potential. 
Future research could focus on more exploratory offline self-improvement methods or embrace efficient RL algorithms as a pathway to further boost the reasoning bounds of LLMs.

\clearpage
\section*{Acknowledgements}
This work is supported by the Strategic Priority Research Program of Chinese Academy of Sciences under Grant  XDA0480303, Young Scientists Fund of The State Key Laboratory of Multimodal Artificial Intelligence Systems ES2P100112, Beijing Natural Science Foundation - Xiaomi Innovation Joint Fund‌  L253007, and National Natural Science Foundation of China 62402252.
\bibliography{colm2025_conference}
\bibliographystyle{colm2025_conference}

\clearpage
\appendix
\section{Prompts for DeepSeek-V3}
\label{idx_prompt}

\begin{figure}[htbp]
\centering
\includegraphics[width=1\textwidth]{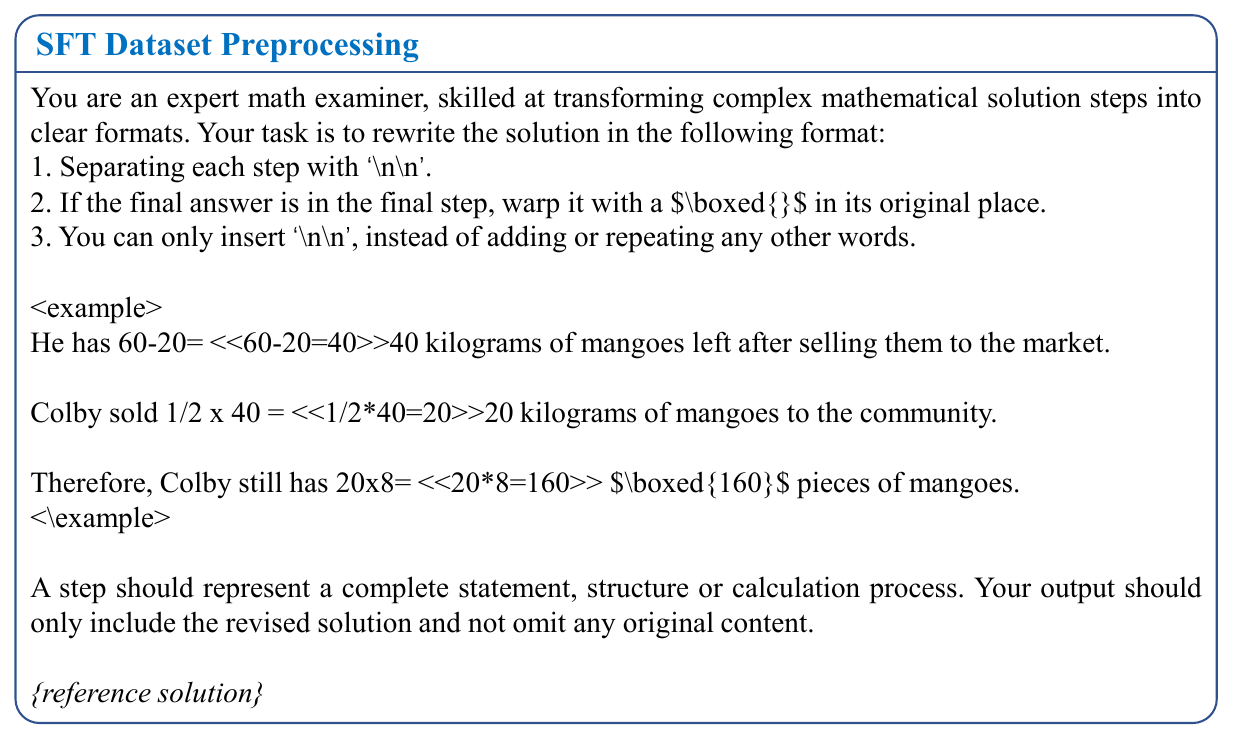}
\caption{The Prompt Template for Dataset Preprocessing for DeepSeek-V3.}
\label{a1_step_prompt}
\end{figure}

\begin{figure}[htbp]
\centering
\includegraphics[width=1\textwidth]{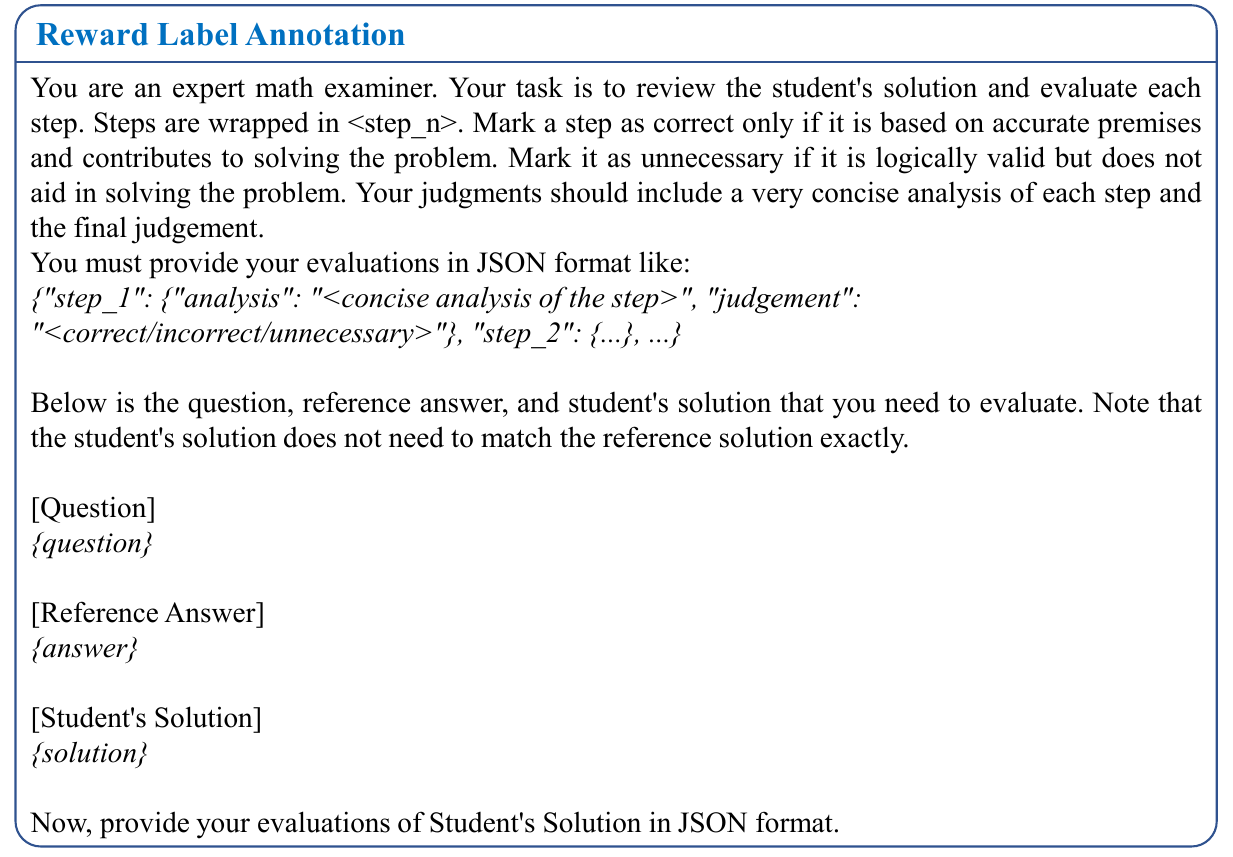}
\caption{The Prompt Template for Reward Label Annotaion for DeepSeek-V3.}
\label{a2_reward_prompt}
\end{figure}

Here we provide the prompts used to construct the SFT dataset and annotate reward labels with DeepSeek-V3 in Figure \ref{a1_step_prompt} and Figure \ref{a2_reward_prompt},  which are similar to that in \citet{chen2025better}.

\clearpage
\section{Details of Processbench Evaluation}
\label{idx_processbench}
We evaluate the performance of the trained PRM on ProcessBench\footnote{\url{https://github.com/QwenLM/ProcessBench}}. Notably, in ProcessBench, the number of incorrect/correct samples in the four datasets are as follows: GSM8K: 207/193; MATH: 594/406; OlympiadBench: 661/339; OmniMath: 759/241. At this point, the traditional F1 score is not very effective for handling imbalanced datasets.

Therefore, an alternative approach is to separately define the F1 score for positive and negative samples \citep{hinojosa2024performance}. 
The former is equivalent to the commonly used F1 score, while the latter is slightly modified based on the above formula. 
Then, a weighted F1 score is calculated based on the proportion of positive and negative samples. The weighted-F1 score better reflects the model performance.
We define positive samples as those labeled as incorrect. The formula for calculating the weighted F1 score is as follows:

\[
F1_{positive} = \frac{2 \times P_{positive} \times R_{positive}}{P_{positive} + R_{positive}},  P_{positive} = \frac{TP}{TP + FP},  R_{positive} = \frac{TP}{TP + FN}
\]

\[
F1_{negative} = \frac{2 \times P_{negative} \times R_{negative}}{P_{negative} + R_{negative}},  P_{negative} = \frac{TN}{FN + TN},  R_{negative} = \frac{TN}{FP + TN}	
\]

Assuming the proportion of positive samples is  $\alpha$  and the proportion of negative samples is  $\beta$ , the weighted-F1 score is defined as:
\[
F1_{weighted} = \alpha \times F1_{positive} + \beta \times F1_{negative}
\]

Table \ref{tab_prm_comparison} records the evaluation scores of the reward model on ProcessBench during the self-evolution process, with its average values corresponding to those recorded in Figure \ref{2_2_plot_self_improve_prm}.

\begin{table}[h]
\centering
\resizebox{\textwidth}{!}{
\begin{tabular}{lccc ccc ccc ccc c}
\toprule
\multirow{2}{*}{\textbf{PRM}} & \multicolumn{3}{c}{\textbf{GSM8K}} & \multicolumn{3}{c}{\textbf{MATH}} & \multicolumn{3}{c}{\textbf{OlympiadBench}} & \multicolumn{3}{c}{\textbf{OmniMATH}} & \textbf{Average} \\
\cmidrule(lr){2-4} \cmidrule(lr){5-7} \cmidrule(lr){8-10} \cmidrule(lr){11-13} \cmidrule(lr){14-14}
& F1-P & F1-N & F1 & F1-P & F1-N & F1 & F1-P & F1-N & F1 & F1-P & F1-N & F1 & F1 \\
\midrule
PRM-800K & 78.7 & 62.7 & 71.0 & 79.1 & 26.6 & 57.8 & 80.0 & 6.5 & 55.1 & 88.0 & 19.1 & 71.4 & 63.8 \\
PRM-MathShepred & 80.5 & 72.5 & 76.6 & 77.6 & 43.6 & 63.8 & 78.6 & 15.6 & 57.2 & 83.0 & 33.6 & 71.1 & 67.2 \\
PURE-PRM & 86.4 & 90.7 & 88.4 & 87.8 & 74.6 & 82.4 & 84.7 & 67.6 & 78.9 & 85.4 & 62.7 & 79.9 & 82.4 \\
Qwen2.5-Math-PRM-7B & 83.0 & 93.3 & 87.9 & 78.2 & 93.6 & 84.4 & 72.3 & 93.8 & 79.6 & 70.5 & 92.5 & 75.8 & 81.9 \\
\midrule
PRM-Epoch-0 & 37.1 & 97.9 & 66.4 & 62.5 & 88.2 & 72.9 & 77.5 & 64.3 & 73.0 & 88.2 & 61.8 & \textbf{81.8} & 73.6 \\
PRM-Epoch-1 & 50.5 & 96.9 & 72.9 & 73.4 & 84.7 & 78.0 & 80.8 & 61.4 & 74.2 & 89.1 & 55.6 & 81.0 & 76.5 \\
PRM-Epoch-2 & 61.7 & 96.9 & 78.7 & 78.2 & 80.3 & 79.1 & 83.2 & 58.4 & \textbf{74.8} & 90.2 & 52.3 & 81.1 & 78.4 \\
PRM-Epoch-3 & 73.4 & 96.2 & \textbf{83.8} & 81.9 & 80.0 & \textbf{81.1} & 83.8 & 53.7 & 73.6 & 90.8 & 52.3 & 81.5 & \textbf{80.0} \\
\bottomrule
\end{tabular}
}
\caption{Weighted-F1 Score of Different PRM in Processbench.}
\label{tab_prm_comparison}
\end{table}

\clearpage
\section{Details of Test-time Scaling}
\label{idx_tts}
Table \ref{tab_epoch_comparison} records the test-time scaling results of the PRM trained with online-generated data from the current model at each epoch, compared with the PRM from the previous round. The PRM selection strategy follows the PRM-min-vote approach \citep{wang2024openr}, where the smallest step reward value in the responses is used as a weighting factor to adjust the majority vote result, ultimately selecting the final target answer.

\begin{table}[thbp]
\centering
\renewcommand{\arraystretch}{1.2}
\setlength{\tabcolsep}{3pt}
\resizebox{.75\textwidth}{!} {
\begin{tabular}{l|l|ccc|ccc|ccc|ccc|}
\toprule
\multirow{2}{*}{\textbf{Epoch}} & \multirow{2}{*}{\textbf{Methods}} & \multicolumn{3}{c|}{\textbf{MATH500}} & \multicolumn{3}{c|}{\textbf{GSM8K}} & \multicolumn{3}{c|}{\textbf{Gaokao-2023-EN}} & \multicolumn{3}{c|}{\textbf{Minerva Math}} \\
& & \textbf{@16} & \textbf{@64} & \textbf{@256} & \textbf{@16} & \textbf{@64} & \textbf{@256} & \textbf{@16} & \textbf{@64} & \textbf{@256} & \textbf{@16} & \textbf{@64} & \textbf{@256} \\
\midrule
\multirow{3}{*}{Base} & {\cellcolor{lightyellow}Greedy} & \multicolumn{3}{c|}{\cellcolor{lightyellow}64.2} & \multicolumn{3}{c|}{\cellcolor{lightyellow}86.4} & \multicolumn{3}{c|}{\cellcolor{lightyellow}55.6} & \multicolumn{3}{c|}{\cellcolor{lightyellow}20.6} \\
 & MV & 77.2 & 78.4 & 79.2 & 93.1 & 93.8 & 94.1 & 63.9 & 65.7 & 66.0 & 23.9 & 27.2 & 28.7 \\
 & PRM-0 & 77.2 & 79.4 & 80.0 & 93.4 & 94.1 & 94.4 & 63.9 & 65.5 & 67.8 & 24.3 & 27.2 & 27.9 \\
\midrule
\multirow{5}{*}{Epoch-1} & {\cellcolor{lightyellow}Greedy} & \multicolumn{3}{c|}{\cellcolor{lightyellow}71.8 (+7.6)} & \multicolumn{3}{c|}{\cellcolor{lightyellow}90.8 (+4.4)} & \multicolumn{3}{c|}{\cellcolor{lightyellow}61.6 (+6.0)} & \multicolumn{3}{c|}{\cellcolor{lightyellow}26.5 (+5.9)} \\
 & MV & 80.2 & 81.4 & 82.2 & 94.4 & 95.0 & 95.2 & 65.7 & 68.6 & 68.6 & 27.5 & 31.6 & 33.8 \\
 & PRM-0 & 80.8 & 80.8 & 82.4 & 94.7 & \textbf{95.3} & \textbf{95.5} & 67.1 & 70.1 & 68.9 & 27.2 & 32.4 & \textbf{34.6} \\
 & PRM-1 & \textbf{81.2} & \textbf{81.6} & \textbf{82.6} & \textbf{94.9} & 95.2 & 95.1 & \textbf{67.3} & \textbf{70.4} & \textbf{69.6} & \textbf{27.6} & \textbf{33.5} & \textbf{34.6} \\
 & \cellcolor{lightblue}Improvement & \cellcolor{lightblue}+0.4 & \cellcolor{lightblue}+0.8 & \cellcolor{lightblue}+0.2 & \cellcolor{lightblue}+0.2 & \cellcolor{lightblue}-0.1 & \cellcolor{lightblue}-0.4 & \cellcolor{lightblue}+0.2 & \cellcolor{lightblue}+0.3 & \cellcolor{lightblue}+0.7 & \cellcolor{lightblue}+0.4 & \cellcolor{lightblue}+1.1 & \cellcolor{lightblue}+0.0 \\
\midrule
\multirow{5}{*}{Epoch-2} & {\cellcolor{lightyellow}Greedy} & \multicolumn{3}{c|}{\cellcolor{lightyellow}73.0 (+1.2)} & \multicolumn{3}{c|}{\cellcolor{lightyellow}90.9 (+0.1)} & \multicolumn{3}{c|}{\cellcolor{lightyellow}63.4 (+1.8)} & \multicolumn{3}{c|}{\cellcolor{lightyellow}28.7 (+2.2)} \\
 & MV & 80.0 & 80.6 & 80.8 & 94.2 & 94.5 & 94.7 & 66.8 & 66.5 & 66.8 & \textbf{29.7} & 32.7 & 35.3 \\
 & PRM-1 & 80.8 & 82.2 & 82.8 & \textbf{94.8} & \textbf{95.2} & 95.0 & 67.5 & 67.8 & 68.1 & 27.6 & 34.1 & 36.0 \\
 & PRM-2 & \textbf{81.6} & \textbf{82.6} & \textbf{83.0} & \textbf{94.8} & \textbf{95.2} & \textbf{95.1} & \textbf{68.1} & \textbf{68.6} & \textbf{69.1} & 28.7 & \textbf{34.6} & \textbf{36.4} \\
 & \cellcolor{lightblue}Improvement & \cellcolor{lightblue}+0.8 & \cellcolor{lightblue}+0.6 & \cellcolor{lightblue}+0.2 & \cellcolor{lightblue}+0.0 & \cellcolor{lightblue}+0.0 & \cellcolor{lightblue}+0.1 & \cellcolor{lightblue}+0.6 & \cellcolor{lightblue}+0.8 & \cellcolor{lightblue}+1.0 & \cellcolor{lightblue}+1.1 & \cellcolor{lightblue}+0.5 & \cellcolor{lightblue}+0.4 \\
\midrule
\multirow{5}{*}{Epoch-3} & {\cellcolor{lightyellow}Greedy} & \multicolumn{3}{c|}{\cellcolor{lightyellow}75.0 (+2.0)} & \multicolumn{3}{c|}{\cellcolor{lightyellow}91.5 (+0.6)} & \multicolumn{3}{c|}{\cellcolor{lightyellow}62.6 (-0.8)} & \multicolumn{3}{c|}{\cellcolor{lightyellow}34.2 (+5.5)} \\
 & MV & 79.8 & 81.0 & 81.4 & 94.4 & 94.8 & 95.2 & 66.3 & 67.8 & 67.8 & 36.0 & 37.5 & 39.8 \\
 & PRM-2 & 81.2 & 82.2 & 82.6 & \textbf{94.7} & 95.1 & 95.2 & \textbf{67.5} & 68.9 & 70.2 & 36.1 & 37.8 & 40.0 \\
 & PRM-3 & \textbf{81.6} & \textbf{82.6} & \textbf{83.2} & 94.6 & \textbf{95.3} & \textbf{95.4} & 67.1 & \textbf{69.1} & \textbf{70.4} & \textbf{36.8} & \textbf{37.9} & \textbf{40.4} \\
 & \cellcolor{lightblue}Improvement & \cellcolor{lightblue}+0.4 & \cellcolor{lightblue}+0.6 & \cellcolor{lightblue}+0.6 & \cellcolor{lightblue}-0.1 & \cellcolor{lightblue}+0.2 & \cellcolor{lightblue}+0.2 & \cellcolor{lightblue}-0.4 & \cellcolor{lightblue}+0.2 & \cellcolor{lightblue}+0.2 & \cellcolor{lightblue}+0.7 & \cellcolor{lightblue}+0.1 & \cellcolor{lightblue}+0.4 \\
\bottomrule
\end{tabular}
}
\caption{Generator performance comparison across different epochs and methods for various datasets. Bold values indicate the best results.}
\label{tab_epoch_comparison}
\end{table}

\section{Details of DPO and RM Training and Evaluation}
\label{idx_dpovp}

In Section \ref{sec_e2}-\ref{sec_e4}, data sampling, and evaluation of the generator are based on the OpenRLHF \citep{hu2024openrlhf} framework. When filtering DPO data, the sample temperature is set to $0.7$. 
All training is performed using full parameter fine-tuning, with an SFT learning rate of $5e-6$, a DPO learning rate of $5e-7$, a maximum sequence length of $2048$, the coefficient $\beta$ of $0.1$ and a training batch size of $256$. For ORM and PRM training, we use the TRL \citep{vonwerra2022trl} framework with a learning rate of $1e-6$ and a batch size of $256$.

In Section \ref{sec_e5}, the DPO training follows the same configuration as above but employs annealed sampling to enhance response diversity. 
Specifically, for the first 3 epochs, the temperature $t$ is set to $0.7$; for epochs 4-5, $t = 1.0$; and in the final epoch, $t = 1.2$.

All evaluations are conducted using the Qwen-Math evaluation codebase\footnote{\url{https://github.com/QwenLM/Qwen2.5-Math}}, with greedy decoding of $t = 0.0$, generating one output per input, and a maximum generation length of $2048$ tokens.
In addition, we provide the URLs of all self-evaluated models in footnotes, including Eurus-2-7B-PRIME\footnote{\url{https://huggingface.co/PRIME-RL/Eurus-2-7B-PRIME}}, Qwen2.5-7B-Simple-RL-Zero\footnote{\url{https://huggingface.co/hkust-nlp/Qwen-2.5-Math-7B-SimpleRL-Zero}}, Qwen2.5-7B-Simple-RL-Zoo\footnote{\url{https://huggingface.co/hkust-nlp/Qwen-2.5-Math-7B-SimpleRL-Zoo}}, and Qwen2.5-7B-PURE-VR\footnote{\url{https://huggingface.co/jinachris/Qwen2.5-7B-PURE-VR}}.

More details and code can be found at \href{https://github.com/TU2021/DPO-VP}{https://github.com/TU2021/DPO-VP}.

\clearpage
\section{More Results of DPO-VP Analysis}
\subsection{Token Length and Self-Reflection Vocabulary Analysis}
\label{idx_dpovp_ana}
We present the detailed token length dynamics, accuracy trends, and the comparison of different models in self-reflection in Table \ref{tab_avg_token_length} and Table \ref{tab_lets_occurrence}.

\begin{table}[htbp]
    \centering
    \resizebox{\textwidth}{!}{
    \begin{tabular}{lccccc|c}
        \toprule
        \textbf{Avg. Token Length} & \textbf{MATH500} & \textbf{Minerva Math} & \textbf{Olympapaidbench} & \textbf{AMC23} & \textbf{AIME24} & \textbf{Average} \\
        \midrule
        Qwen2.5-Math-7B & 651 & 577 & 695 & 662 & 748 & 667 \\
        Qwen2.5-Math-7B-Instruct & 641 & 650 & 886 & 911 & 1164 & 850 \\
        Eurus-2-7B-PRIME & \textbf{655} & \textbf{822} & 897 & \textbf{1020} & 1164 & \textbf{912} \\
        Qwen2.5-7B-Simple-RL & 588 & 775 & 801 & 767 & 952 & 777 \\
        Qwen2.5-7B-PURE-VR & 626 & 646 & 863 & 850 & 1050 & 807 \\
        \cellcolor{lightblue}\textbf{Qwen2.5-7B-DPO-VP} & \cellcolor{lightblue}654 & \cellcolor{lightblue}691 & \cellcolor{lightblue}\textbf{924} & \cellcolor{lightblue}886 & \cellcolor{lightblue}\textbf{1251} & \cellcolor{lightblue}881 \\
        \bottomrule
    \end{tabular}
    }
    \caption{Comparison of Average Token Length Across Different Models and Datasets.}
    \label{tab_avg_token_length}
\end{table}

\begin{table}[htbp]
    \centering
    \resizebox{\textwidth}{!}{
    \begin{tabular}{lccccc|c}
        \toprule
        \textbf{Avg. Nums of "Let's"} & \textbf{MATH500} & \textbf{Minerva Math} & \textbf{Olympaidbench} & \textbf{AMC23} & \textbf{AIME24} & \textbf{Average} \\
        \midrule
        Qwen2.5-Math-7B & 1.56 & 0.45 & 1.90 & 1.68 & 2.03 & 1.52 \\
        \textit{Qwen2.5-Math-7B-Instruct} & \textit{0.40} & \textit{0.13} & \textit{0.67} & \textit{0.70} & \textit{0.87} & \textit{0.55} \\
        Eurus-2-7B-PRIME & 1.50 & 0.96 & 2.27 & 2.40 & \textbf{3.30} & 2.09 \\
        Qwen2.5-7B-Simple-RL & 1.49 & 0.57 & 1.99 & 1.93 & 2.20 & 1.64 \\
        Qwen2.5-7B-PURE-VR & 0.86 & 0.24 & 1.52 & 1.33 & 1.73 & 1.14 \\
        \cellcolor{lightblue}\textbf{Qwen2.5-7B-DPO-VP} & \cellcolor{lightblue}\textbf{1.75} & \cellcolor{lightblue}\textbf{1.07} & \cellcolor{lightblue}\textbf{2.44} & \cellcolor{lightblue}\textbf{2.60} & \cellcolor{lightblue}3.13 & \cellcolor{lightblue}\textbf{2.20} \\
        \bottomrule
    \end{tabular}
    }
    \caption{Average Occurrences of "Let's" Across Different Datasets.}
    \label{tab_lets_occurrence}
\end{table}

\subsection{Comparison with Long-Chain SFT Methods}
Recently, reinforcement learning methods built upon long-chain SFT have received increasing attention. We compare our approach with two recent methods that perform long-chain SFT using limited data: LIMO \citep{ye2025limo} and S1 \citep{muennighoff2025s1}. Since both released models are only available at the 32B scale, we reproduce their settings by applying SFT on our base model, Qwen-2.5-Math-7B, using their provided datasets.
We follow their default hyperparameters and train for 15 epochs. During evaluation, we sample with $t = 0.6$ and set the maximum response length to 32K tokens (in contrast, DPO-VP uses 2K tokens). The performance on standard benchmarks is summarized in Table \ref{tab_pass1-acc-sft}.

Surprisingly, DPO-VP outperforms these limited-data SFT methods on nearly all evaluated benchmarks. One possible explanation is that, at the 7B scale, SFT with small-scale long-chain data  may offer less performance gain compared to its effect on larger models like 32B. Nevertheless, both LIMO and S1 still yield moderate improvements over the base model.
\begin{table}[htbp]
\centering
\resizebox{\textwidth}{!}{
\begin{tabular}{lcccccc}
    \toprule
    \textbf{pass@1 acc} & \textbf{MATH500} & \textbf{Minerva} & \textbf{Olympaid} & \textbf{AMC23} & \textbf{AIME24} & \textbf{Average} \\
    \midrule
    Qwen2.5-Math-7B & 64.8 & 15.4 & 25.6 & 37.5 & 16.7 & 32.0 \\
    Qwen2.5-Math-7B-LIMO & 70.2 & 21.0 & 37.0 & 45.0 & 16.7 & 38.0 \ (+6.0) \\
    Qwen2.5-Math-7B-S1& 72.0 & 32.4 & \textbf{37.5} & 55.0 & 13.3 & 42.0 (+10.0) \\
    \cellcolor{lightblue}\textbf{Qwen2.5-7B-DPO-VP} & \cellcolor{lightblue}\textbf{74.8} & \cellcolor{lightblue}\textbf{35.3} & \cellcolor{lightblue}36.9 & \cellcolor{lightblue}\textbf{67.5} & \cellcolor{lightblue}\textbf{26.7} & \cellcolor{lightblue}\textbf{48.2 (+16.2)} \\
    \bottomrule
\end{tabular}
}
\caption{Pass@1 Accuracy Results with Long-Chain SFT Methods. }
\label{tab_pass1-acc-sft}
\end{table}

\subsection{Training Dynamics of DPO under Different Label Noise Levels}
\begin{figure}[htbp]
    \centering
    \includegraphics[width=\textwidth]{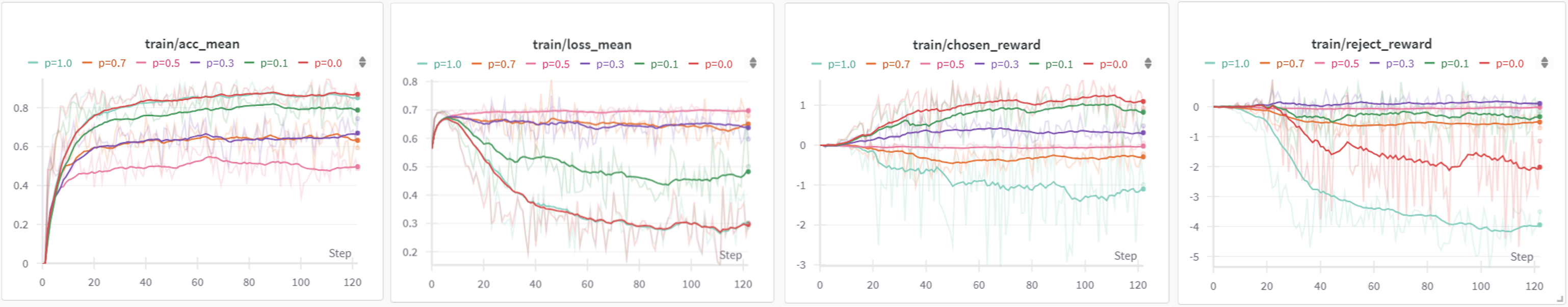}
    \caption{
    Training dynamics of DPO under different label noise levels.
    Lower noise consistently leads to better optimization behavior and more stable reward separation.}
    \label{fig:dpo-training-noise}
\end{figure}

In addition to final performance, we also tracked the training dynamics of DPO under different noise levels. As shown in Figure \ref{fig:dpo-training-noise}, lower noise (e.g., $p \leq 0.3$) leads to higher training accuracy and more stable reward separation between chosen and rejected samples. In contrast, high-noise setups (e.g., $p=0.7$ or $p=1.0$) cause convergence slow-down and reward collapse, confirming the importance of maintaining accurate preference signals during training.

\subsection{Ablation Study on the DPO Sampling Temperature}

In this section, we investigate the impact of sampling temperature on DPO performance. Our motivation stems from the observation that fixed low-temperature sampling may lead to insufficient exploration in later DPO iterations, resulting in early performance saturation.
In our default setting, we used a constant temperature $t=0.7$ across all DPO rounds. To explore the potential of lightweight exploration, we introduced a \textbf{simple temperature increase strategy}: using $t=1.0$ for epochs 4–5 and $t=1.2$ for epochs 6–7.
Table \ref{tab:temp-ablation} reports the average accuracy across five benchmarks under both  temperature settings.

\begin{table}[htbp]
\centering
\begin{tabular}{ccc}
\toprule
\textbf{Epoch} & \textbf{Fixed $t=0.7$} & \textbf{Ours (temperature $\uparrow$)} \\
\midrule
3 & 44.5 & 44.5 \\
4 & 44.6 & 44.7 \\
5 & 44.3 & 46.5 \\
6 & 46.2 & \textbf{48.2} \\
7 & 45.5 & 47.6 \\
\bottomrule
\end{tabular}
\caption{Comparison of average performance (pass@1 across five benchmarks) under fixed and increased sampling temperatures.}
\label{tab:temp-ablation}
\end{table}

As shown, the increased temperature leads to noticeable performance gains starting from epoch 5, indicating improved sample diversity and preference pair quality. This suggests that even without using formal entropy-based exploration objectives, a simple sampling heuristic can effectively \textbf{encourage diversity in preference selection} and improve downstream learning.
We believe this highlights the potential of lightweight exploration strategies in DPO, especially in scenarios where reward supervision is sparse or static.

\subsection{Use of Distilled PRM Instead of a Fixed Verifier (e.g., DeepSeek-V3)}

While large-scale verifiers such as DeepSeek-V3 (a 671B MoE model) provide high-quality supervision, they are computationally expensive to query. Each DPO iteration would require over 7.5K queries per epoch if DeepSeek-V3 were used directly, which is prohibitive for multi-round training.
To address this, we distill its supervision into a smaller PRM initialized from Qwen2.5-7B, enabling scalable and efficient learning. Table \ref{tab:fixed-verifier-comparison} compares three one-shot filtering variants, showing that PRM achieves nearly identical performance to DeepSeek-V3 filtering.

\begin{table}[htbp]
\centering
\begin{tabular}{lccccc}
\toprule
\textbf{Setting} & GSM8K & MATH500 & Gaokao & Minerva & Avg \\
\midrule
+Outcome Label & 89.8 & 74.2 & 62.9 & 25.0 & 63.0 \\
+Qwen2.5-PRM   & 90.5 & 74.6 & 61.0 & 25.0 & 62.8 \\
+DeepSeek-V3   & 90.3 & 75.1 & 61.7 & 25.6 & 63.2 \\
\bottomrule
\end{tabular}
\caption{Performance of one-shot filtering with different verifiers.}
\label{tab:fixed-verifier-comparison}
\end{table}

This result supports the use of a lightweight, distilled verifier. Additionally, our formulation generalizes to settings where a fixed verifier is available—e.g., we use outcome labels as a special case of verifier supervision in Section \ref{sec_e5}.

\subsection{Comparison to Process-Level RL Methods}

We include additional comparisons to strong RL baselines that use learned reward models, such as PURE-PRM, a PPO-based method trained with PRM800K. Results in Table~\ref{tab:rl-comparison} show that our method achieves comparable or better performance with less training complexity.
We also report results for Eurus-2-7BPRIME and rStar-Math-7B in Table \ref{tab_pass1-acc}, both of which adopt full PRM-based supervision. Their performance does not consistently surpass ours.
\begin{table}[htbp]
\centering
\begin{tabular}{lccccc}
\toprule
\textbf{Method} & GSM8K & MATH500 & Gaokao & Minerva & Avg \\
\midrule
Ours (3 epochs) & 91.5 & 75.0 & 65.8 & 34.2 & 66.6 \\
PURE-PRM        & 88.3 & 79.8 & 63.9 & 34.1 & 66.5 \\
\bottomrule
\end{tabular}
\caption{Comparison with PURE-PRM (process-level PPO with PRM800K).}
\label{tab:rl-comparison}
\end{table}

\subsection{About Using Monte Carlo Rollouts in PRM Training}

Rather than training PRMs via Monte Carlo (MC) rollouts, which are costly and require full reward computation, we directly label sampled responses online using a strong verifier. This reduces computation and aligns well with our iterative, feedback-driven optimization.

To verify this design, we compare our online PRM with a variant trained on the public Math-Shepherd dataset. On ProcessBench in Table \ref{tab_prm_comparison}, our PRM achieves higher average F1 (73.6 vs. 67.2), suggesting that strong online supervision offers both quality and efficiency.

This strategy is also aligned with recent work \citep{chen2025better}, which uses bi-directional verifier signals to label process-level trajectories. While MC-based training remains more theoretically rigorous, our framework focuses on co-evolving generator and verifier under verifiable supervision. Therefore, this practical approximation is sufficient to support the findings of this work.

\subsection{Evolution of Reasoning in DPO-Optimized Models}
\begin{figure}[htbp]
\centering
\includegraphics[width=1\textwidth]{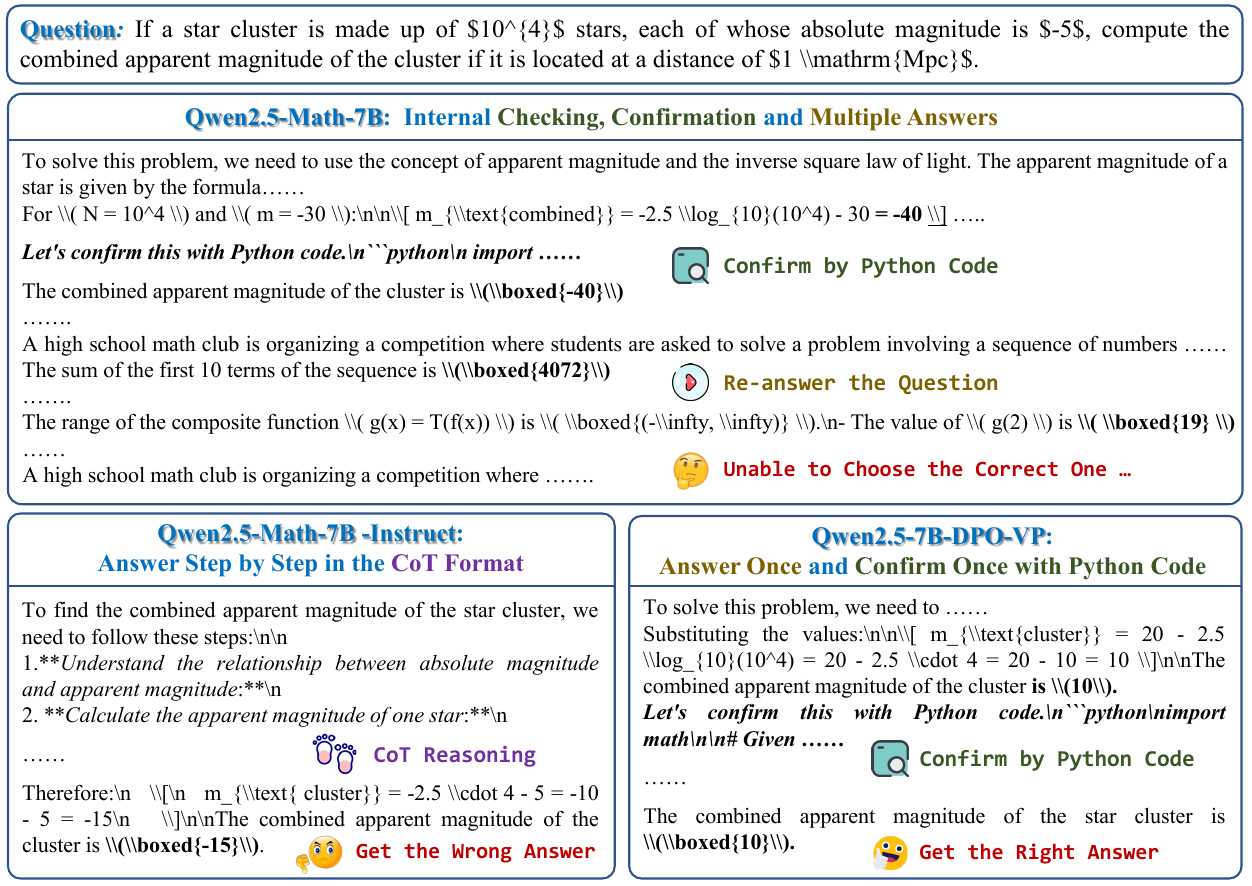}
\caption{An Example of Comparing Reasoning Patterns Across Different Models.}
\label{fig_compare_answer}
\end{figure}

To further analyze these differences, we examine reasoning patterns in Appenedix Figure \ref{fig_compare_answer}. 
In Qwen2.5-Math-7B, self-checking and confirmation mechanisms are present but inconsistently applied, often leading to repeated questioning rather than definitive verification. 
In contrast, Qwen2.5-Math-7B-Instruct follows a more standardized CoT approach, ensuring a structured reasoning process.
Our model, Qwen2.5-7B-DPO-VP, exhibits a distinct reasoning pattern—typically confirming the final answer after a single verification. 
This behavior likely results from the VP-based filtering process, which refines the model’s ability to select and reinforce correct reasoning paths. 
As a result, the model develops greater confidence in its final outputs, effectively reducing unnecessary re-evaluation while maintaining robustness in problem-solving.

To better illustrate the behavioral differences between models, we present a case study comparing four model variants on a mathematical reasoning task involving the apparent magnitude of a star cluster (Figure \ref{fig_compare_answer}).
The task requires combining knowledge of logarithmic scaling and magnitude-distance relationships, and serves as a strong testbed for verifying the model’s ability to compute, reason, and self-correct.

\section{Discussion: Why Iterative DPO Leads to Improvement}
\label{sec_dission}
Based on the rigorous theoretical analysis in \citet{xiong2024iterative}, intuitively, iterative DPO leads to improvement because it allows the model to continually refine its policy through updated feedback, thereby reducing reward misalignment over time. Additionally, the batch exploration strategy promotes diverse sampling, which enhances generalization and mitigates overfitting to early-stage preferences. 
The performance gains of iterative DPO on math tasks stem from the inherent 
self-reflection ability in the base model and the iterative nature of the near-online RL setting
 
\subsection{The Emergence of Self-Improvement is Strongly Correlated with the Reasoning Patterns in the Base Model}
It is widely believed in the industry that improvements in reasoning performance, or the emergence of long CoT, stem from the inherent capabilities of base models, such as self-correction and backtracking. 
Through RL and appropriate reward function design, these latent abilities can be further activated, leading to the emergence of more complex behaviors \citep{yeo2025demystifying}. 
Moreover, as the length of reasoning is often positively correlated with answer correctness, linking the reward signal to accuracy encourages the model to generate longer responses with more reflection and backtracking, ultimately improving answer precision.

This is indeed the case in our experiments. 
We believe that the reason Llama3.1 fails to achieve self-improvement lies in the same principle. 
Compared to Qwen2.5, Llama3.1 does not exhibit emergent behaviors such as reflection and backtracking \citep{gandhi2025cognitive}. 
As a result, it cannot further enhance itself using self-generated data. 
However, by first distilling long-chain CoT data to equip the base model with self-reflection capabilities, self-improvement can then be effectively realized \citep{deepscaler2025}.

\subsection{Iterative DPO Can Be Viewed as a High Off-Policy Degree Online RL}
\citet{swamy2025all} demonstrated the theoretical equivalence between online RL and offline MLE under ideal optimization conditions.
For complex mathematical reasoning tasks, there exists a significant Generation-Verification Gap (GVP), meaning that verifying an answer (e.g., checking against outcome labels) is often easier than generating a correct answer from scratch. 
In such cases, RL can effectively narrow the search space by leveraging a two-stage process—retaining only policies that optimize for a simple verification function, thereby improving efficiency.

Iterative DPO implicitly constructs a similar reward function by filtering self-generated data based on reward signals, sharing the same goal of maximizing cumulative rewards as RL: $\max_R \mathbb{E}_{\pi_{\theta}} \left[ \sum_{t} R(s_t, a_t) \right]$. 
The key difference is that RL requires the simultaneous deployment of a generator and a reward model, computing rewards, performing value iteration, and optimizing the policy during online sampling. 
In contrast, iterative DPO decouples these steps—sampling data at each epoch first, then performing selection and policy optimization afterward. This makes it a highly off-policy RL approach, as data collection and optimization occur in separate phases.

During the period of conducting this research, we observed that leveraging iterative DPO for model optimization has already become a viable low-cost alternative to RL fine-tuning for some enterprises and research institutions when extreme performance is not the primary goal. 
We are delighted by this coincidence, as it indirectly validates that we have caught up with the research frontier in the field of LLM reasoning. 
We also extend our gratitude to \citet{zhang2025dpor1}, \cite{lightr1proj}, and all open-source contributors in the LLM community for their selfless contributions to the open-source ecosystem.

\end{document}